\RequirePackage{fix-cm}
\documentclass[smallcondensed]{svjour3}     
\usepackage[numbers,sort]{natbib}
\makeatletter
\renewcommand*{\@biblabel}[1]{\hfill[#1]}
\makeatother

\smartqed  
\usepackage{graphicx}
\usepackage{float}
\usepackage{subfig}
\usepackage{multirow}
\usepackage{natbib}
\usepackage[misc]{ifsym}

\newcommand{\citeup}[1]{\textsuperscript{\cite{#1}}}
\begin{document}

\title{Unified Regularity Measures for Sample-wise Learning and Generalization}


\author{Chi Zhang\and
	Xiaoning Ma \and
	Yu Liu\and 
	Le Wang\and 
	Yuanqi Su\and
	Yuehu Liu
}


\institute{Chi Zhang$^{1,2,}$\textsuperscript{\Letter}, 
	Xiaoning Ma$^{3}$, 
	Yu Liu$^{1,2}$, 
	Le Wang$^{1,2}$, 
	Yuanqi Su$^{3}$, 
	Yuehu Liu$^{1,2}$      
	\at
	\textsuperscript{\Letter} Chi Zhang is the corresponding author. \email{colorzc@stu.xjtu.edu.cn} 
	\at
	$^{1}$ Institute of Artificial Intelligence and Robotics, Xi'an Jiaotong University, Xi'an, China. 
	\at
	$^{2}$ College of Artificial Intelligence, Xi'an Jiaotong University, Xi'an, China. 
	\at
	$^{3}$ School of Computer Science and Technology, Xi'an Jiaotong University, Xi'an, China.
}

\date{Received: date / Accepted: date}

\maketitle

\begin{abstract}
Fundamental machine learning theory shows that different samples contribute unequally both in learning and testing processes. Contemporary studies on DNN imply that such sample difference is rooted on the distribution of intrinsic pattern information, namely \textit{sample regularity}. Motivated by the recent discovery on network memorization and generalization, we proposed a pair of sample regularity measures for both processes with a formulation-consistent representation. Specifically, cumulative binary training/generalizing loss (CBTL/CBGL), the cumulative number of correct classifications of the training/testing sample within training stage, is proposed to quantize the stability in memorization-generalization process; while forgetting/mal-generalizing events, i.e., the mis-classification of previously learned or generalized sample, are utilized to represent the uncertainty of sample regularity with respect to optimization dynamics. Experiments validated the effectiveness and robustness of the proposed approaches for mini-batch SGD optimization. Further applications on training/testing sample selection show the proposed measures sharing the unified computing procedure could benefit for both tasks.
\end{abstract}

\section{Introduction}
\label{intro}
Deep learning has been widely applied and made great progress in many fields such as image recognition and synthesis, video understanding, and natural language processing in recent years. However, the generalization of neural networks is still affected by the unexplainable black-box structure of algorithms, biased datasets, noisy labeling, and various evaluation metrics, becoming an open problem. In particular, the over-parameterized deep neural network has far more parameters than  training samples. Although, it is predicted to severely overfit by classical learning theory, it generalizes remarkably well\citeup{overfit}. In this regard, Zhang\citeup{rethinking} found that the deep neural network trained using SGD can easily fit random labels. In this case, Rademacher complexity, which is a measurement of the ability to fit random noise in the statistical learning theory, will be approximately equal to 1. This will result in loose and invalid generalization error bounds. A similar reasoning is also appropriate for VC-dimension\citeup{rethinking}. It shows that the traditional learning theory fails to explain why deep neural networks generalize well from the training set to new data.

Recently, some researchers have tried to develop a new generalization theory adapted to over-parameterized deep neural networks by explaining the phenomenon formation mechanism from the memorization - generalization theory of deep learning. Zhang\citeup{rethinking} argues that there are remaining signals in correctly labeled training data and neural networks can not only capture them but also rely on memory to forcibly fit noisy part. Therefore, the ability to extract patterns while memorizing exception samples is the real reason for the high performance generalization of deep neural networks. Feldman\citeup{feldman} further pointed out that the memorization of rare and atypical samples is the source of unreasonable beyond-expectation generalization performance\citeup{beyond} of deep neural networks due to the long-tailed distribution of images in real world. Both of them imply differences in the distribution of patterns contained in samples and the powerful memory of the neural networks. A fundamental question then arises: \textbf{\textit{How to represent the distribution of intrinsic pattern in samples?}}

Differences in sample distribution have become one of the common bottlenecks in the field of deep learning, such as training set biases, i.e., inconsistency in the distribution between training and testing sets. Existing machine learning researches show that the amount of information contained in samples and their contributions to training are not equal due to difference in distribution\citeup{notallequal}. Reweighing training samples and training in order of difficulty can help alleviate sample biases. OHEM\citeup{OHEM} represents the distribution based on loss and selects difficult samples; Data Dropout\citeup{dropout} thinks that the impact of samples on training can be quantified by the influence function\citeup{understandingblackbox}; Iou-balanced sampling\citeup{Ioubalancedsampling} explores the sample distribution on detection tasks and proposes an Iou-based representation space in which training data are collected uniformly to improve generalization performance; GHM\citeup{GHM} argues that sample differences can be reflected by gradient norm; PISA\citeup{PISA} proposes a novel IoU hierarchical local rank strategy as a quantitative way to evaluate sample differences. More recently, a new line of research has been opened, which empirically represent sample distribution based on a memorization - generalization theory. Jiang\citeup{cscore} describes that cumulative binary training loss (CBTL) can be used as a lightweight proxy for samples' consistency score; Toneva\citeup{forgettingevents} novelly proposes a forgetting event based sample representation.

Inspired by the above work, we attempt to explore the way of representing distribution of intrinsic pattern in samples through the memorization - generalization theory, empirically demonstrating the exposed regularity of the underlying patterns in the data. However, we find that unidimensional sample representation with CBTL or the number of forgetting events alone can cause confusion in distinguishing samples. Therefore, we incorporate long-term stable information and short-term dynamic signals during training to propose unified regularity measures with bi-dimensional representation. They represent the distribution of training samples and testing samples in a CBTL - forgetting events and a CBGL - mal-generalizing events space, respectively. This study will be beneficial in the fields of few-shot learning, training acceleration, difficult sample selecting, data distribution consistency, algorithm testing and so on. The major contributions of this paper are summarized as:
\begin{itemize}
	\item[$\bullet$] A bi-dimensional representation involves forgetting events and CBTL is proposed to measure the regularity of training samples.
	\item[$\bullet$] Likewise, a bi-dimensional representation combining newly defined CBGL and mal-generalizing events, is proposed to measure the regularity of testing samples.
	\item[$\bullet$] Our experimental findings suggest that samples with higher regularity seem to contribute little in both training and testing task, which in return validates the effectiveness of the proposed measures.
\end{itemize}

\section{Preliminary Works}
\label{sec:1}
The general form\citeup{lihang} of the learning system can be described as the follows:

Given a training dataset: $T=\left\lbrace \left( x_{1},y_{1}\right), \left( x_{2},y_{2}\right),\cdots,\left( x_{N},y_{N}\right) \right\rbrace$, where $\left( x_{i},y_{i}\right),$ $i=1,2,\cdots,N$ denotes the input observation - label sample pair. The learning system is trained with the given training dataset to obtain a model, denoted as a conditional probability distribution $\hat{P}\left(Y|X\right)$ or a decision function $Y=\hat{f}\left(X\right)$, to describe the mapping between input and output random variables. The optimal model is generally trained by the strategy of minimizing the empirical risk $R_{emp}\left(\hat{f}\right)=\frac{1}{N}\sum_{i=1}^{N}L(y_{i},\hat{f}(x_{i}))$, where $L(\cdot)$ is the loss.
\subsection{Forgetting Events and Cumulative Binary Training Loss}
\label{sec:2}
Continuous learning in real world requires that intelligent systems are able to learn on successive tasks without performance degradation on the preceding training tasks, just as humans do. Researchers have found that this problem setting poses a great challenge for connectionist-based neural networks\citeup{french,mccloskey,ratcliff}. The phenomenon, known as "catastrophic forgetting", is mainly manifested by the tendency of neural networks to quickly and brutally forget all the acquired knowledge of the previous task (e.g., task A) after the current task (e.g., task B) is added. Even deep neural networks, which have achieved great successes in recent years, are not able to overcome this shortcoming well\citeup{kirkpatrick}, which undoubtedly increases the doubts about the research on general artificial intelligence based on them. 
The main source of this phenomenon is that when the task goal changes, the acquired weights of the network adapted to the previous task also adjust to the needs of the new task, thus failing to generalize on the previous task.

Inspired by this, Toneva\citeup{forgettingevents} argues that a single learning task optimization based on mini-batch stochastic gradient descent can be considered as a process similar to continuous learning. In this process, each mini-batch of training data can be considered as a small task that is sequentially handed over to the deep neural network. This leads to the following definition of sample forgetting events\citeup{forgettingevents}.
\paragraph{\textbf{Forgetting Events}}: During the mini-batch sample learning process, the acquired (i.e., correctly classified) training sample at time $t$ are misclassified at subsequent time $t'$ ($t'>t$).

Toneva\citeup{forgettingevents} explored the memory dynamics during training and classified the samples into forgettable and unforgettable samples based on forgetting events. At the same time, they experimentally verified the features atypicality and visual illegibility of the forgettable samples. However, both the extremely quickly learned simple samples and difficult learned exception samples have few forgetting events, and apparently the statistics of forgetting events are symmetric. They cannot well distinguish the differences in samples. In response, Jiang\citeup{cscore} quantified the regularity of samples by measuring the consistency of the sample with the overall distribution of all samples through a lightweight proxy of CBTL, defined as follows.
\paragraph{\textbf{Cumulative Binary Training Loss (CBTL)}}: During the mini-batch sample learning process, the cumulative number of correct classifications of the training sample up to time $t$.

\begin{figure}
	\centering
	\includegraphics{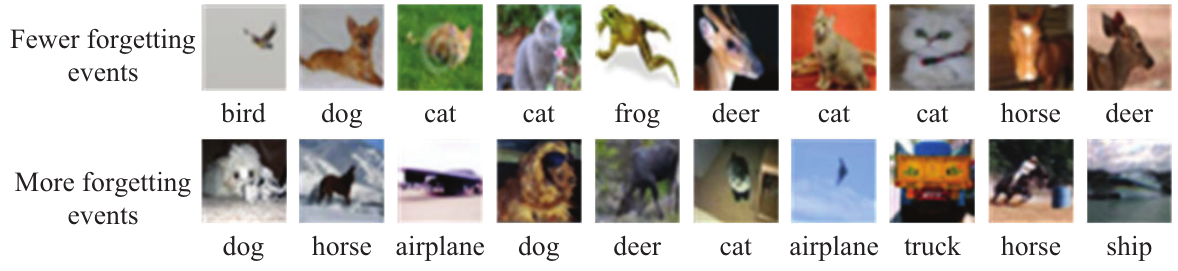}
	\caption{Examples in Cifar-10 training set. Each column has the same CBTL, with fewer forgetting events in the top row than in the bottom row. Obviously, the top samples are more regular and recognizable. The visual difficulties of samples in the same column are different, so using only CBTL to distinguish pattern differences in samples can cause confusion.}
	\label{1}
\end{figure}

The ResNet-110 network was trained with Cifar-10 dataset and CBTL and the number of forgetting events were recorded. Figure \ref{1} reflects that samples in the Cifar-10 training set are not equally difficult to classify when CBTL is the same but the number of forgetting events is different. It is obvious that the images in the top row are more regular and easier to recognize than the bottom row. At this point,  CBTL alone can't distinguish the pattern differences in samples well.

Inspired by this, we try to represent differences in the pattern distribution of samples based on forgetting events jointly with CBTL. We consider CBTL as a long-term stability measure of the learning difficulty of one sample. The occurrence of a forgetting event, on the other hand, implies that the sample crossed in the wrong direction as the decision boundary changed, demonstrating that the direction of the loss incurred by the sample at this moment is not consistent with the overall loss of the training set. Therefore, statistics of forgetting events can be considered as a measure of uncertainty embedded in the sample in learning a decision boundary.

\begin{figure*}
	\centering
	\includegraphics[width=1\textwidth]{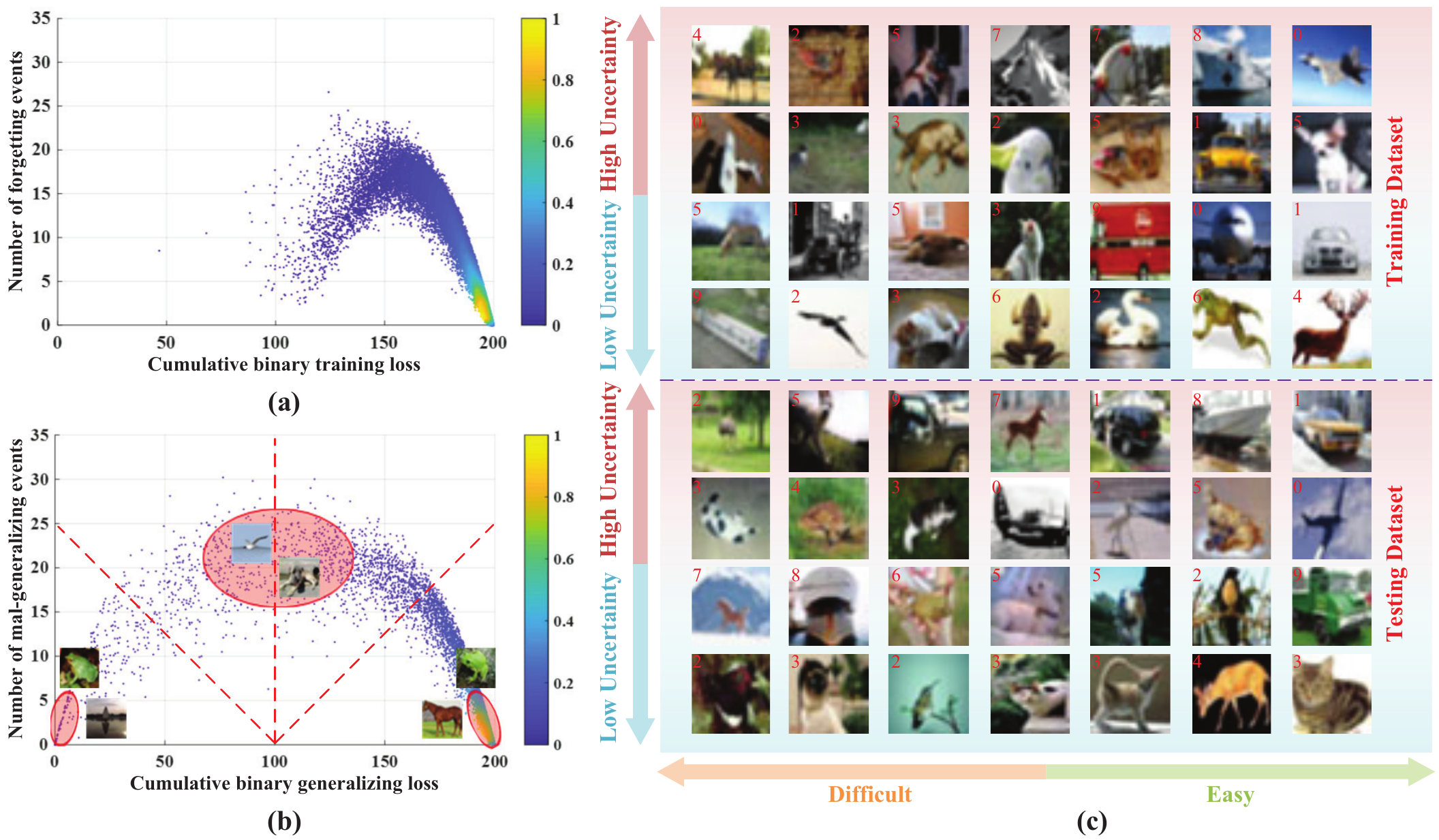}
	\caption{Representation of samples' distribution in Cifar-10 for (a) training set, (b) testing set and (c) visualization. (a) and (b) are distribution scatter plots, where each point represents a sample, and the right bar is color-coded to distinguish the density intensity. The sample density is the number of samples per unit area in its neighborhood, calculated by the number of samples in the circle with this sample as the center and a certain value $r$ as the radius divided by the area of the circle. Here the radius is tentatively set to 6.8, which is calculated by $r=\sqrt{(\frac{x}{30})^2+(\frac{y}{30})^2}$, where x and y denote the range of the horizontal and vertical axis of the representation, respectively. The top half of (c) is the training samples and the bottom half is the testing samples. In (c), the horizontal axis represents binary training/generalizing loss, which is easier the further to the right, and the vertical axis represents forgetting/mal-generalizing events, with higher uncertainty the further up. The numbers in the upper left corner of the image represent category numbers: 0 - airplane, 1 - automobile, 2 - bird, 3 - cat, 4 - deer, 5 - dog, 6 - frog, 7 - horse, 8 - ship, and 9 - truck.}
	\label{2}
\end{figure*}

As a result, in this paper, we empirically propose a bi-dimensional sample regularity representation, which is in the CBTL - forgetting events and CBGL - mal-generalizing events spaces, as shown in Figure \ref{2}. Some good properties of this representation are explored and the insight of dataset compressibility is validated in the later section.

Due to high computational cost of performing statistics after each mini-batch during training, this paper takes the update after each epoch, as detailed in the following Experimental Setup section. From this, we formalize the above definitions of forgetting events and CBTL as follows.

The predicted label for the training example $x_{i}$ obtained after $t$ epochs of optimization is denoted as $\hat{y}_{i}^{t}=\hat{f}^{t}(x_{i})$ . Let $acc^{t}_{i}=1_{\hat{y}_{i}^{t}=y_{i}}$ , which is a binary variable, indicating whether the sample is correctly classified at time epoch $t$. The formalized definitions are as follows.

\paragraph{\textbf{Forgetting Events}}: Let $for_{i}^{t}=1_{acc^{t-1}_{i}=1,acc^{t}_{i}=0}$, the number of forgetting events of one training sample $(x_{i},y_{i})$ at epoch $t$ is defined as follows:
\begin{equation}
F^{t}_{i}=\sum_{n=1}^{t}for_{i}^{n} 
\label{e1}
\end{equation} %

\paragraph{\textbf{Cumulative Binary Training Loss (CBTL)}}: For training sample $(x_{i},y_{i})$,  CBTL at epoch $t$ is defined as follows:
\begin{equation}
L^{t}_{i}=\sum_{n=1}^{t}acc_{i}^{n} 
\label{e2}
\end{equation} %
\subsection{Mal-generalizing Events and Cumulative Binary Generalizing Loss}
\label{sec:3}
This paper is equally concerned with the dynamics of generalization on the testing set during sequential learning of mini-batch data. Imitating definitions of forgetting events and CBTL, the following definitions are available.

\paragraph{\textbf{Mal-generalizing Events}}:  During the mini-batch sample learning process, the testing samples can be correctly classified at epoch $t$ but misclassified at subsequent epoch $t'$ ($t'>t$).

\paragraph{\textbf{Cumulative Binary Generalizing Loss (CBGL)}}: During the mini-batch sample learning process, the cumulative number of correct classifications of testing samples up to epoch $t$.

The formal description of the above definitions is similar to equation \ref{e1} and \ref{e2}, with the only difference being applied to the samples in the testing dataset.

\section{Experimental Verification and Analysis}
\subsection{Experimental Setup}
\label{sec:4}
As described in Toneva's work\citeup{forgettingevents}, it would be quite time and computationally expensive to calculate whether a forgetting event occurs for all training samples after each mini-batch. Therefore they only calculate for the mini-batch samples involved in the training after each mini-batch. The calculation of mal-generalizing events faces the same dilemma, i.e., it is not feasible to calculate the generalization status of all testing samples after each mini-batch. Then, considering the limited impact on model performance after mini-batch samples training, this paper adopts a very different strategy from the above approach, i.e., updating the inference states of all training and testing samples after each epoch. To confirm this idea, the results of our strategy is compared with that proposed by Toneva's\citeup{forgettingevents}. We trained the ResNet-110 network with Cifar-10 dataset. The average errors of CBTL and the number of forgetting events are 1.2614 and 0.6762, respectively. The Pearson correlation coefficient of the vectors of CBTL and the number of forgetting events for 50000 samples in Cifar-10 training set are 0.9896 and 0.9835, respectively, both of which are very strongly correlated. Therefore, the strategy of making one inference and updating the states after one epoch is used as a lightweight proxy. When the number of training samples is 50,000, the number of testing samples is 10,000, and the batch size is 128, the number of model inferences required is reduced to about 1/390 of the ideal case. Note that this approximation is likely to aggravate the randomness of model generalization.

This paper mainly explores inference states based on the training process of ResNet-110 with the Cifar-10 dataset. The model is trained for a total of 200 epochs, and its average training performance is close to the highest accuracy of the architecture on the Cifar-10 dataset, i.e., $93.53\%$. In particular, the initial learning rate is 0.1, which decreases to 0.01 at the 81st epoch and to 0.001 at the 122nd epoch. In this paper, the same network is trained 10 times under the same hyper-parameter settings and its mean value is taken to eliminate the effect of randomness of model inference on the empirical analysis.
\subsection{Representation of Sample Distribution}
\label{sec:5}
The sample distribution is represented from the perspective of  memorization - generalization theory of neural networks. Inspired by Jiang's\citeup{cscore} and Toneva's\citeup{forgettingevents} works, this paper empirically proposes a bi-dimensional sample regularity representation, which is in the CBTL - forgetting events and CBGL - mal-generalizing events spaces. This section explores this representation.

\subsubsection{Unidimensional Sample Distribution Representation}
\begin{figure}[!t]
	\centering
	\subfloat[CBTL]{\includegraphics[width=0.45\textwidth]{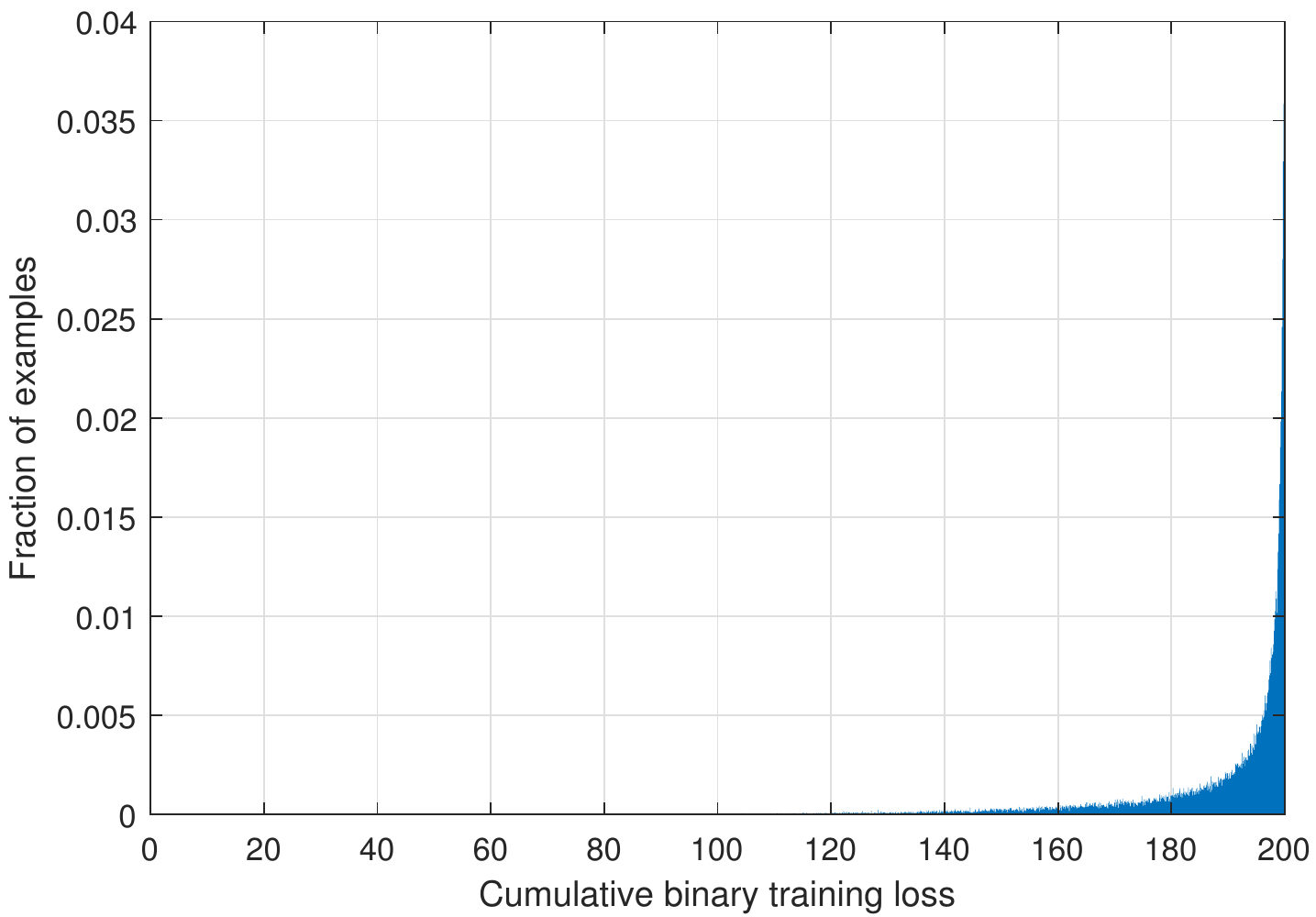}%
		\label{3a}}
	\hfil
	\subfloat[Forgetting events]{\includegraphics[width=0.45\textwidth]{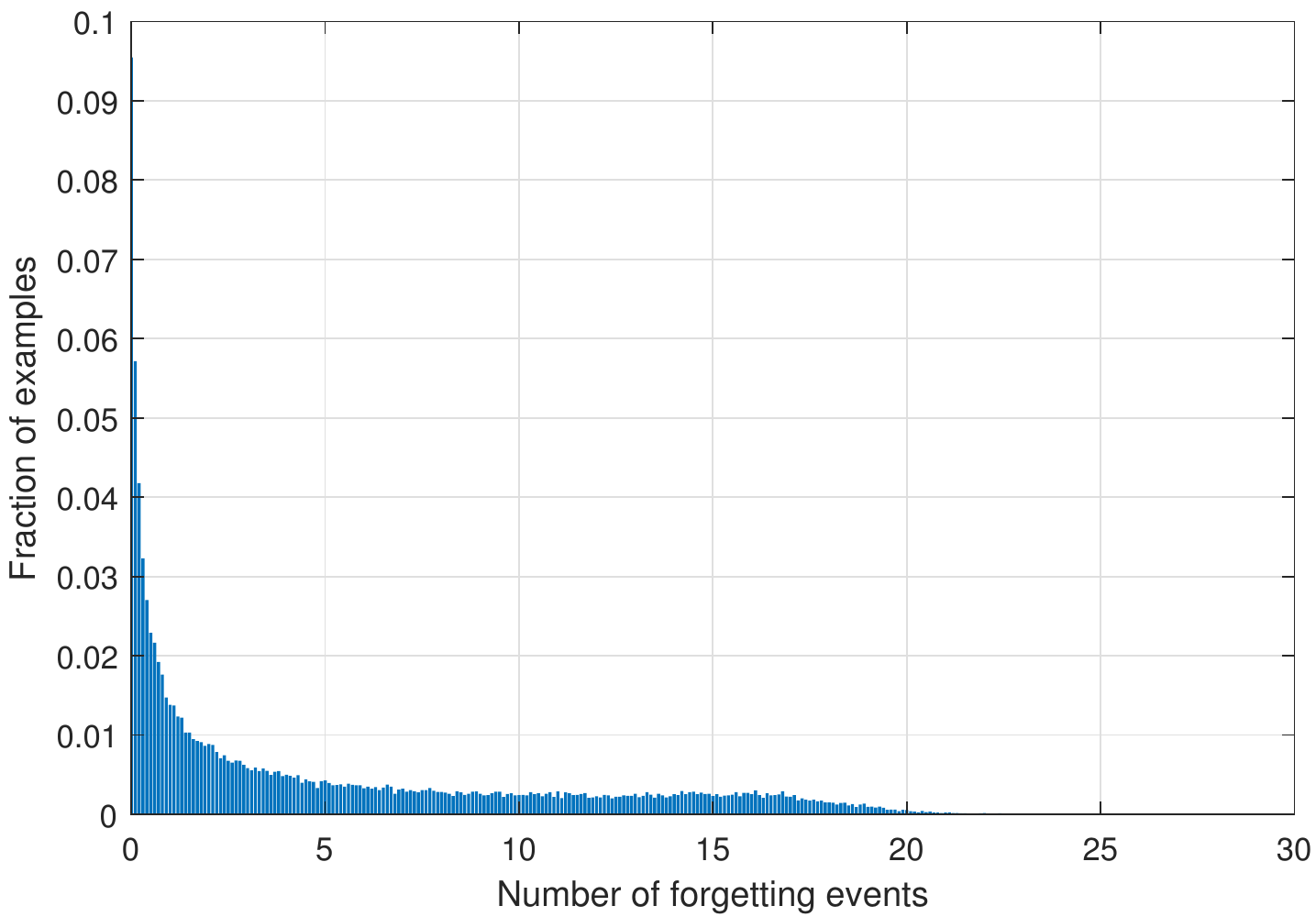}%
		\label{3b}}
	\hfil
	\subfloat[CBGL]{\includegraphics[width=0.45\textwidth]{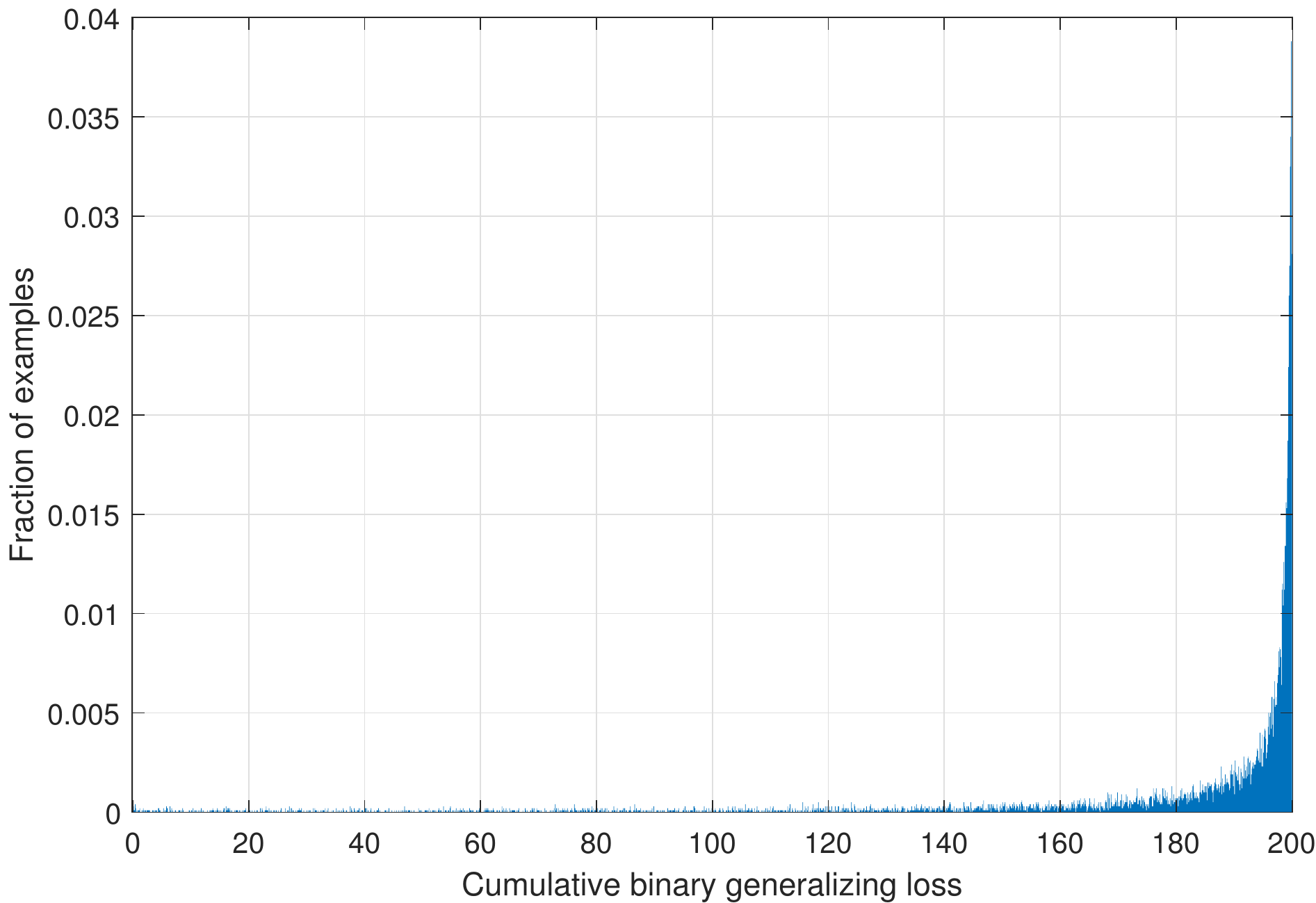}%
		\label{3c}}
	\hfil
	\subfloat[Mal-generalizing events]{\includegraphics[width=0.45\textwidth]{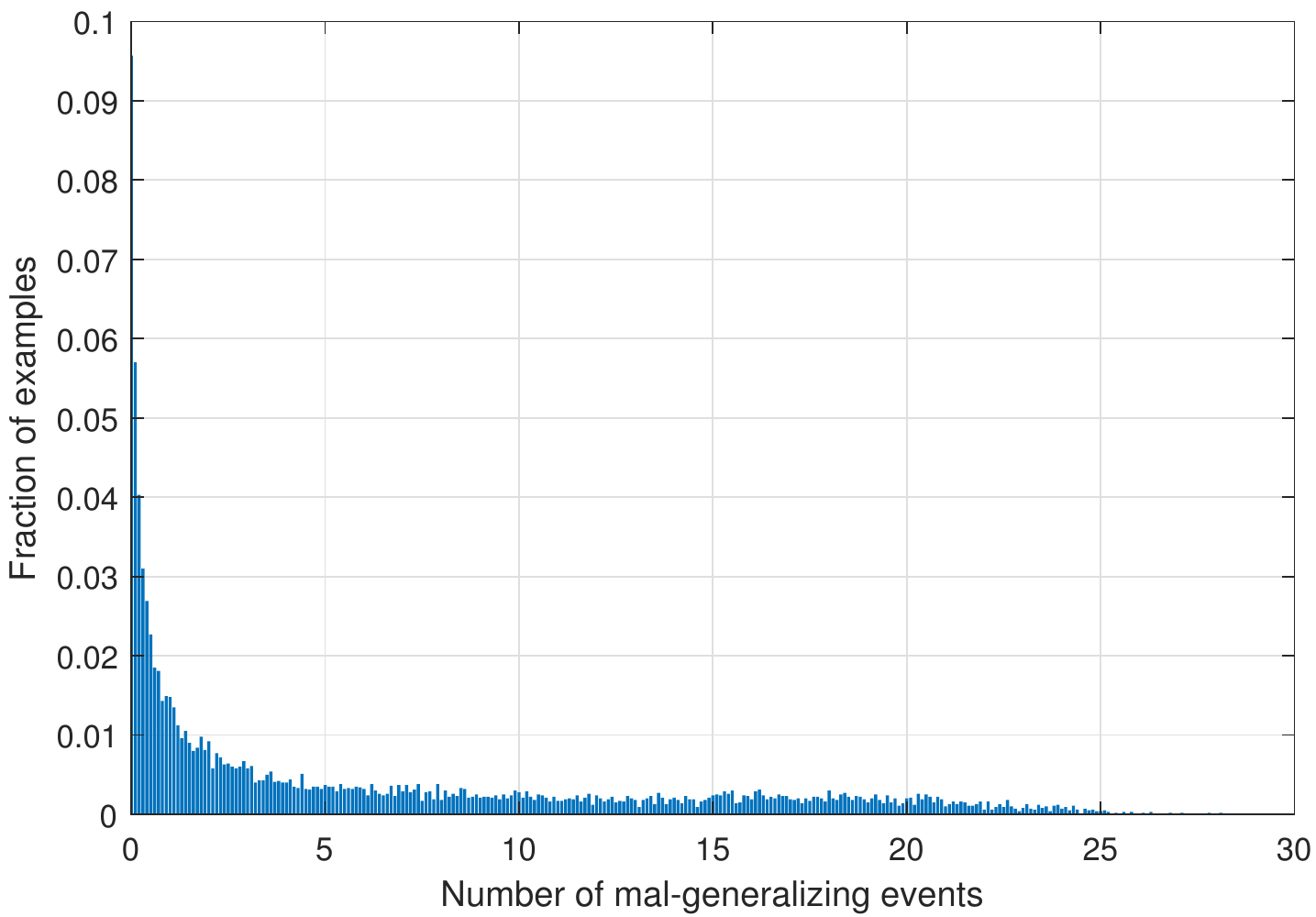}%
		\label{3d}}
	\hfil
	\caption{Histogram of the sample distribution in a single dimension. The distributions are all long-tailed and similar for the testing and training samples. The distributions in forgetting events and mal-generalizing events have smaller range and more detailed local information.}
	\label{3}
\end{figure}

Firstly, the histogram of the sample distribution in a single dimension is made as Figure \ref{3}.

The distributions in Figure \ref{3} are all long-tailed, with simple samples having high CBTL/CBGL and a small number of forgetting/mal-generalizing events dominating. The distribution of training and testing samples is very similar, but the testing samples are not involved in training, resulting in a longer tail. Forgetting/Mal-generalizing events have a smaller range of the distribution and more detailed local information than CBTL/CBGL.

\subsubsection{Bi-dimensional Sample Distribution Representation}
\begin{figure}[!t]
	\centering
	\subfloat[Training samples]{\includegraphics[width=0.4\textwidth]{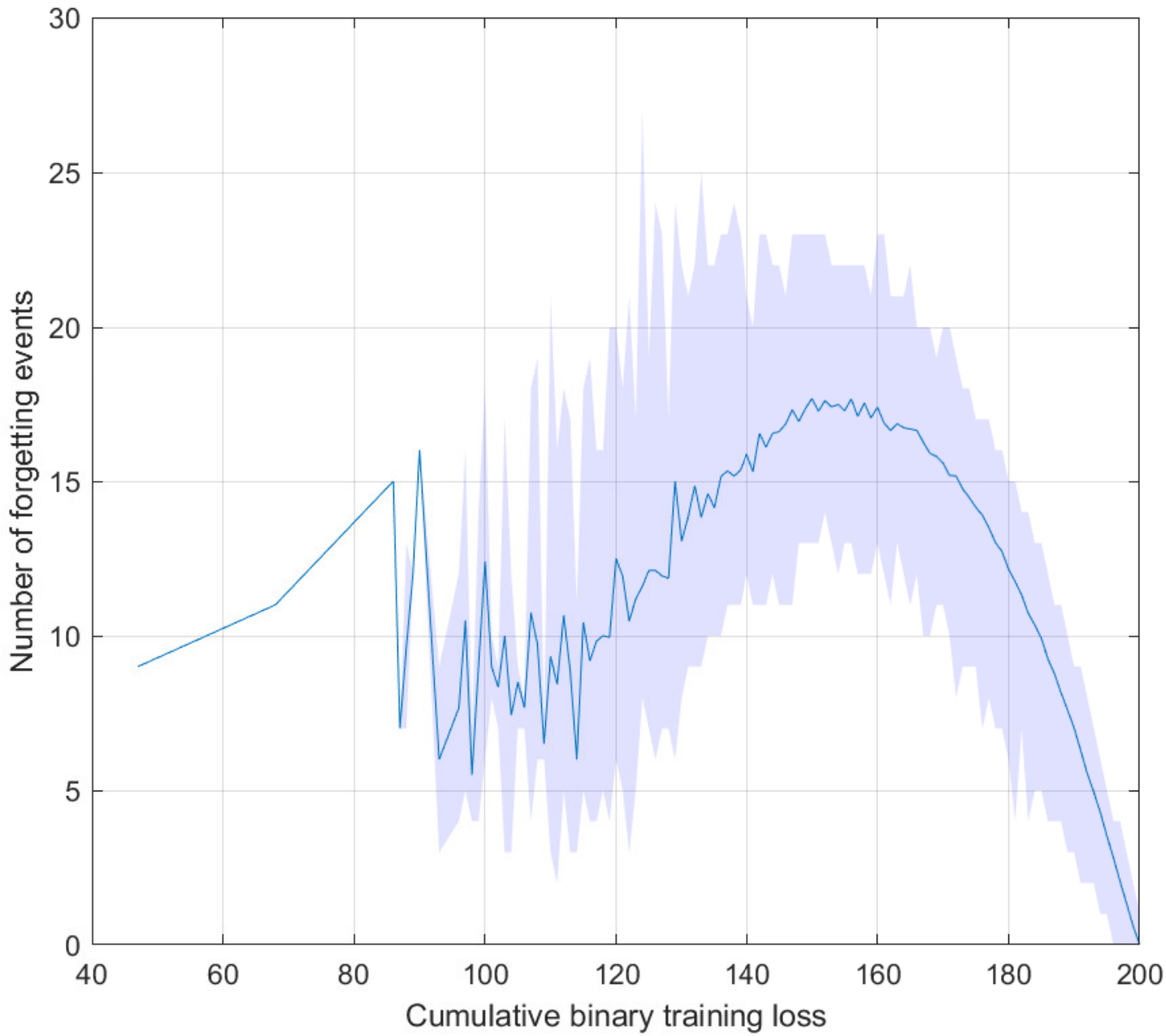}%
		\label{4a}}
	\hfil
	\subfloat[Testing samples]{\includegraphics[width=0.4\textwidth]{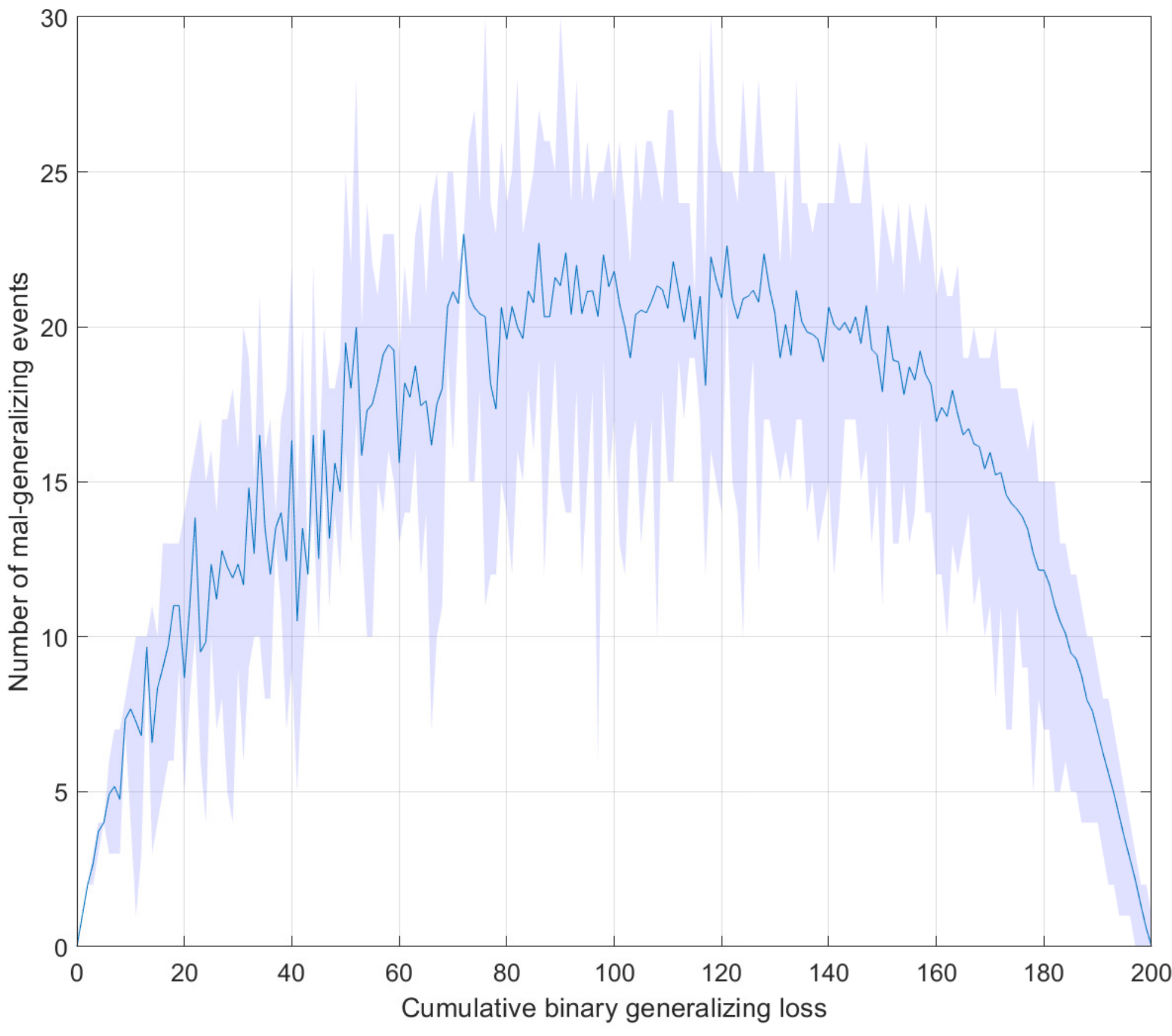}%
		\label{4b}}
	\hfil
	\caption{The number of forgetting/mal-generalizing events under the same binary training/ generalizing loss for the training/testing samples, where the blue line shows the mean value and the blue transparent band represents the fluctuation range.}
	\label{4}
\end{figure}

Further, the regularity of samples' distribution is depicted in the CBTL/CBGL - forgetting/mal-generalizing events space as shown in Figure \ref{2} (a) and (b). The samples are symmetrically distributed in our defined bi-dimensional space with respect to the forgetting/mal-generalizing events, while the distribution of forgetting/mal-generalizing events under the same binary training/testing loss is widely disparate, as shown in Figure \ref{4}. This likewise confirms the idea that a single dimension does not distinguish well the intrinsic patterns of the samples.

From Figure \ref{2} (a) and (b), they also show that the sample density in the lower right corner is larger, i.e., the number of simple samples is dominant, which is consistent with the previous findings. The symmetry distributions of the testing and training samples are also approximate, but the symmetry center axis of the training samples is more to the right and the distribution is tighter. It is also consistent with the intuitive understanding that the empirical error tends to be smaller than the test error.

\subsubsection{Visual Verification}
We visualized some samples at different positions of the distribution as shown in Figure \ref{2} (b). It can be seen that the pattern distribution of the samples does differ. For the samples in the lower right corner, the brown horse on the green grass and the green frog on the gray rock, the objects are highly recognizable and regular. The samples in the middle part, on the other hand, have some confusion between the background and foreground objects, which are more difficult to identify. For the samples in the bottom left corner, the cat with green light and the triangular boat, both with unconventional category features, they look more like a frog and a tree, and are very easy to be misclassified.

\subsection{Exploring the Measures' Properties}
\label{sec:6}
\subsubsection{Training Randomness} 
Due to the randomness of the optimization process and the uncertainty of model inference, the results of repetitive experiments with the same hyper-parameter settings for the same network architecture vary, so the stability needs to be explored. In this paper, we explore the statistics of 10 repetitive experiments.

\begin{figure*}[!t]
	\centering
	\subfloat[Testing \#1]{
		\begin{minipage}{0.23\textwidth}
			\includegraphics[width=1\textwidth]{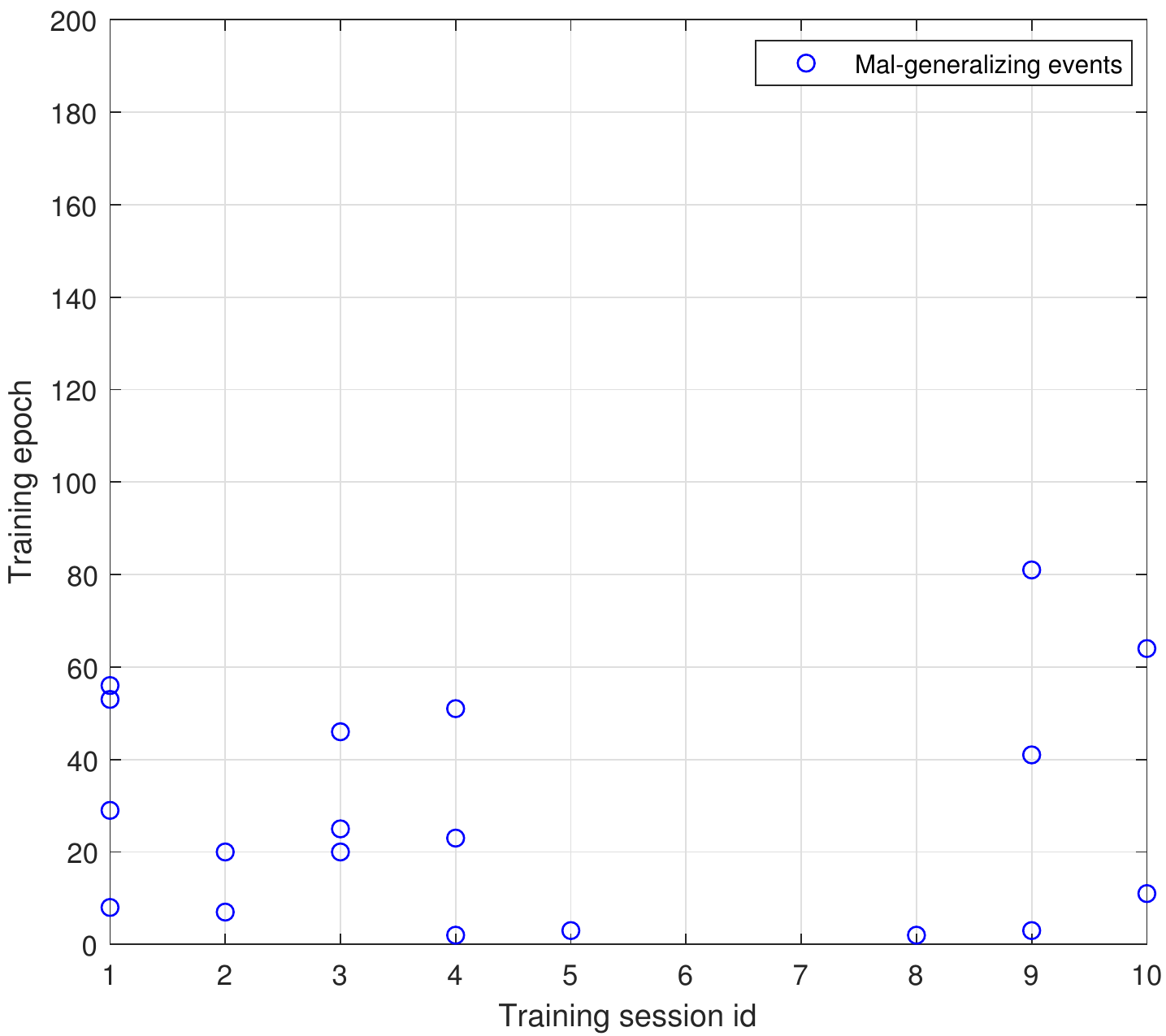} \\
			\includegraphics[width=1\textwidth]{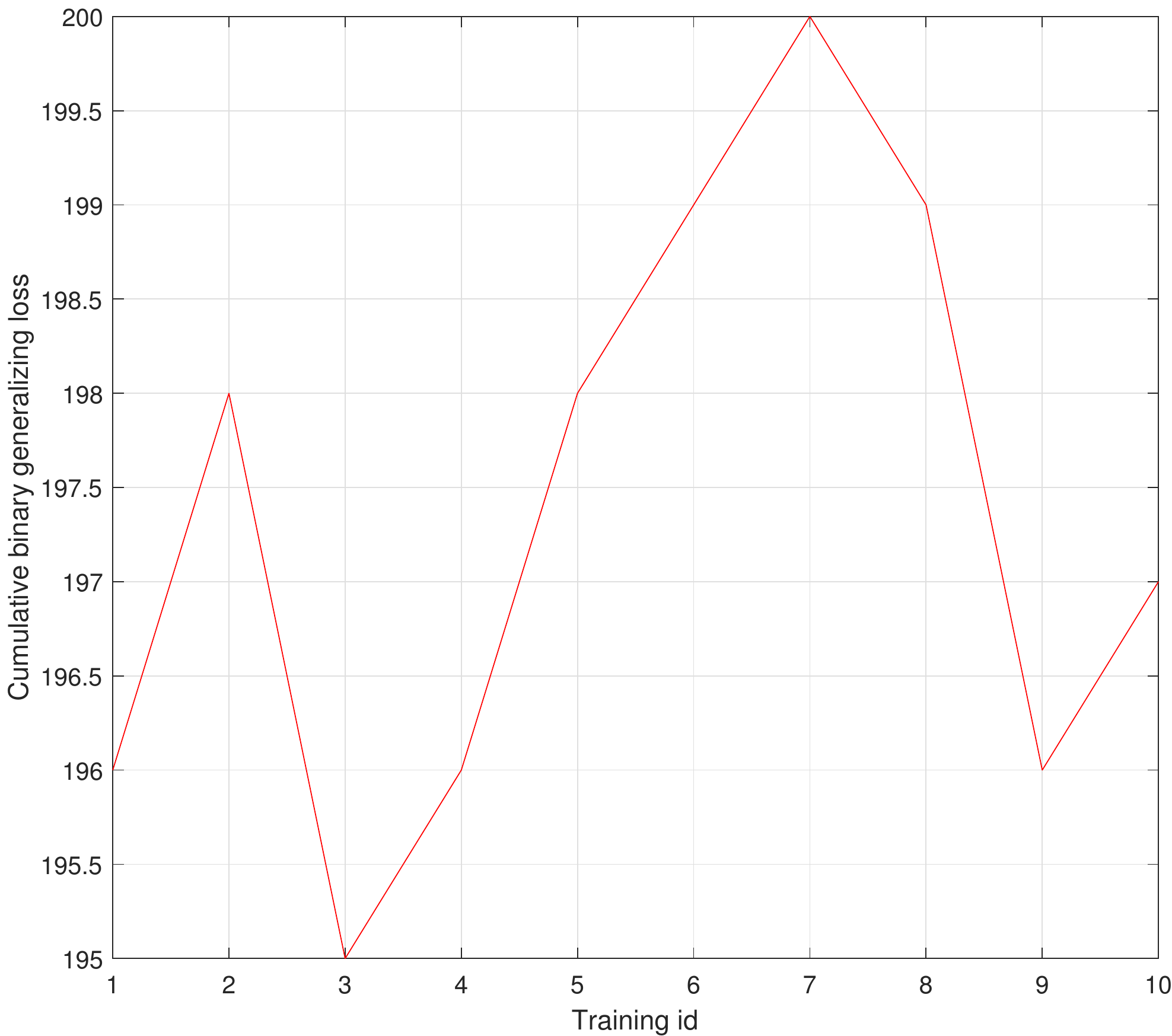}
		\end{minipage}
		\label{5a}
	}
	\subfloat[Testing \#3]{
		\begin{minipage}{0.23\textwidth}
			\includegraphics[width=1\textwidth]{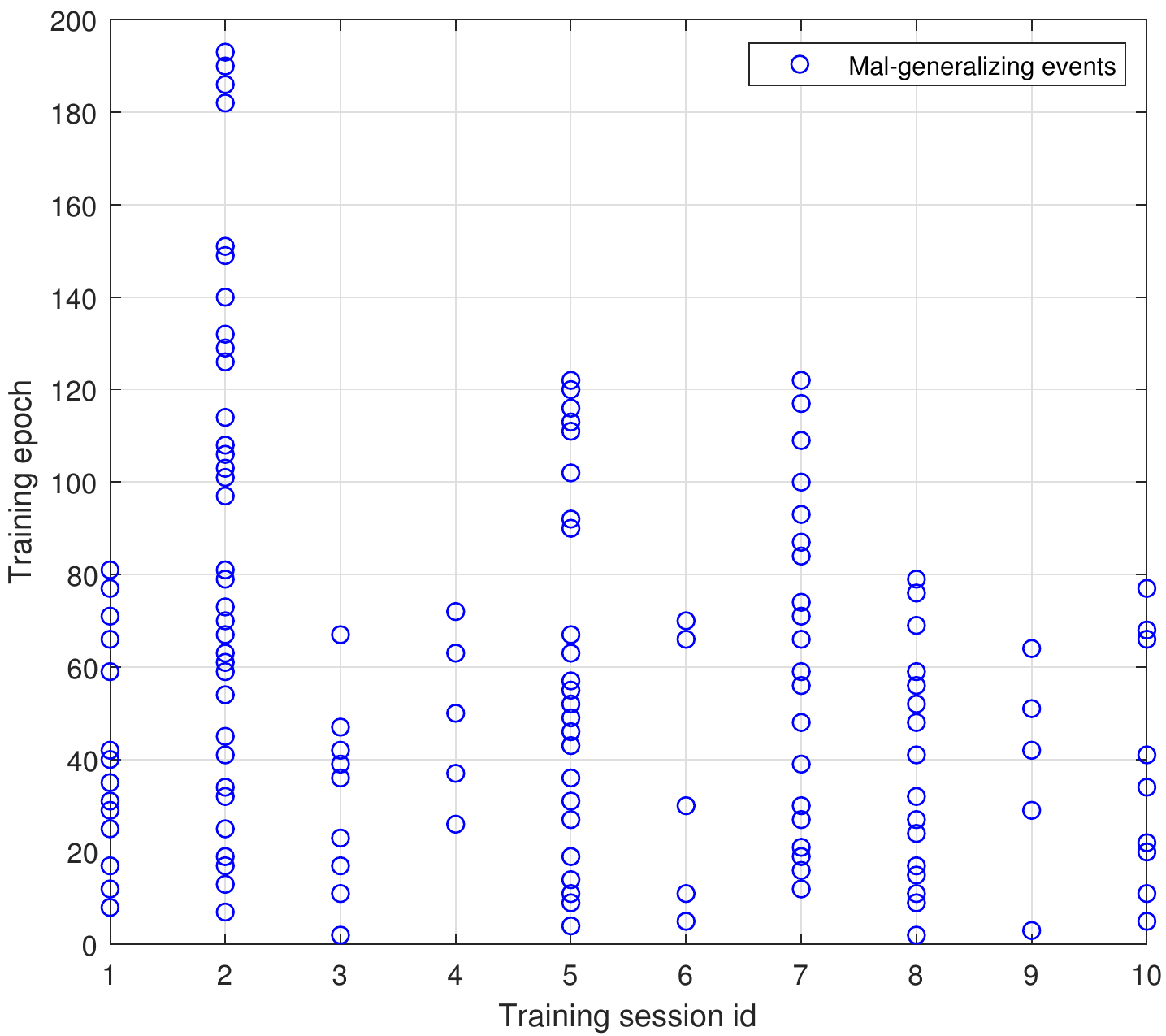} \\
			\includegraphics[width=1\textwidth]{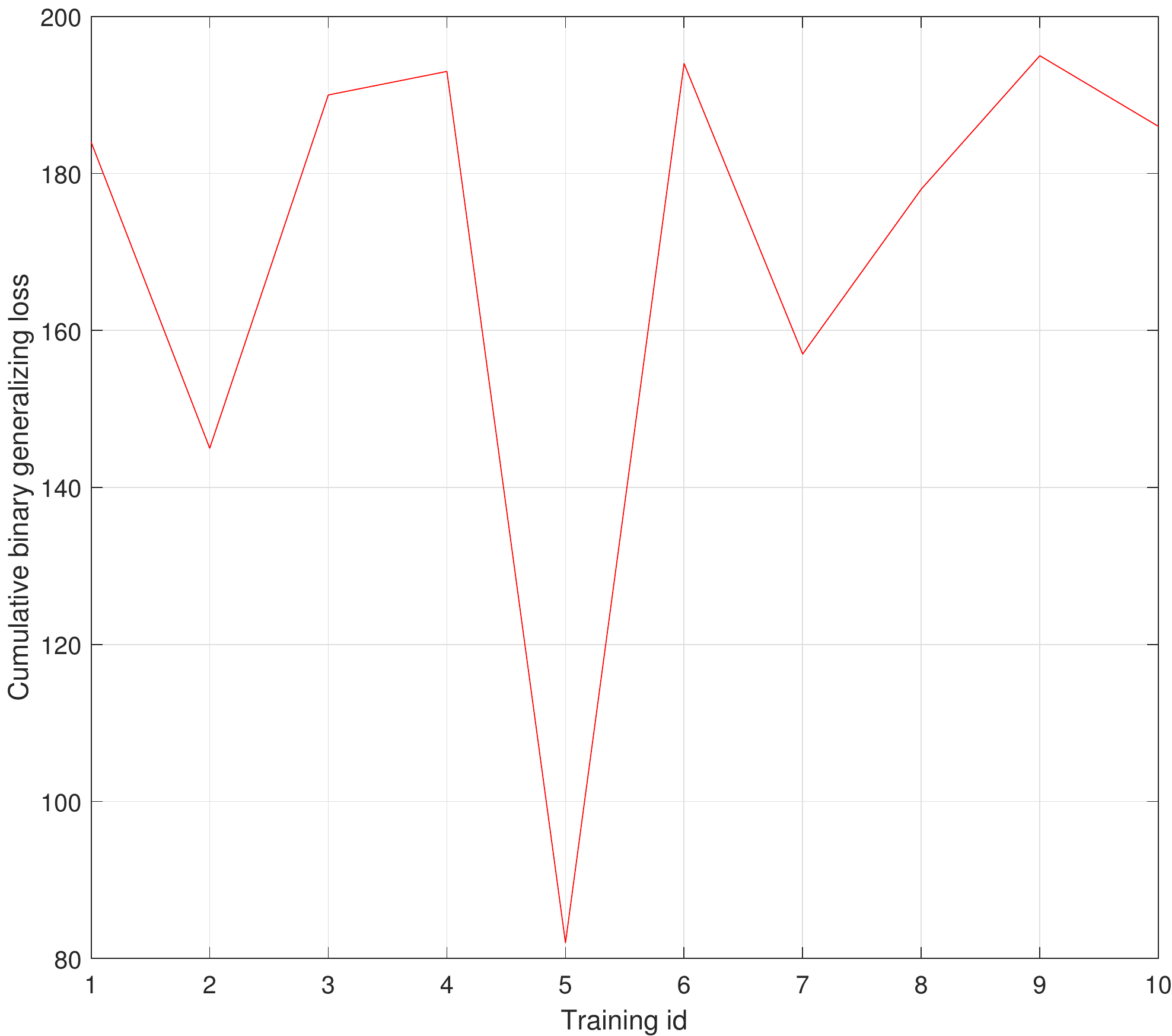}
		\end{minipage}
		\label{5b}
	}
	\subfloat[Training \#7]{
		\begin{minipage}{0.23\textwidth}
			\includegraphics[width=1\textwidth]{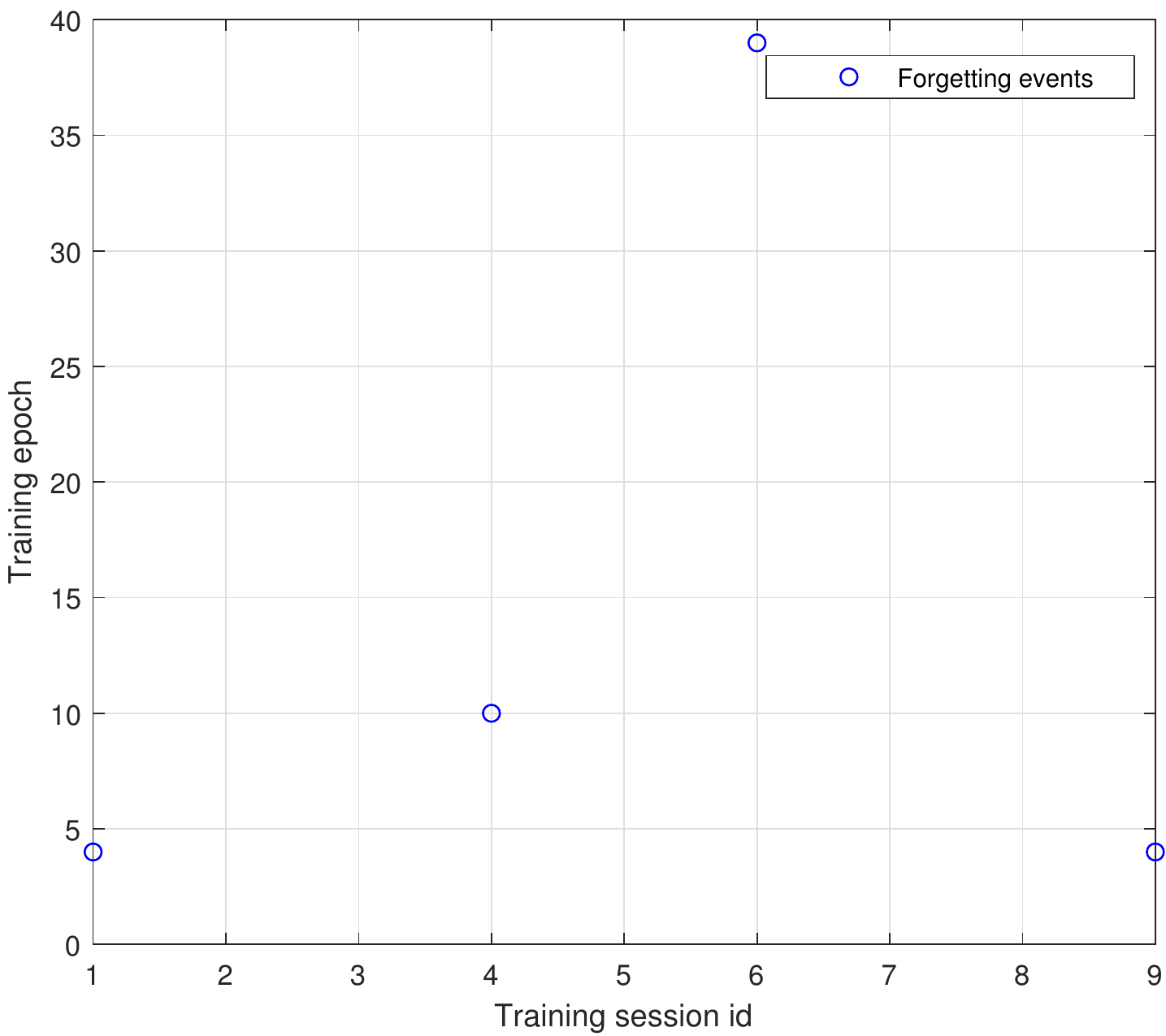} \\
			\includegraphics[width=1\textwidth]{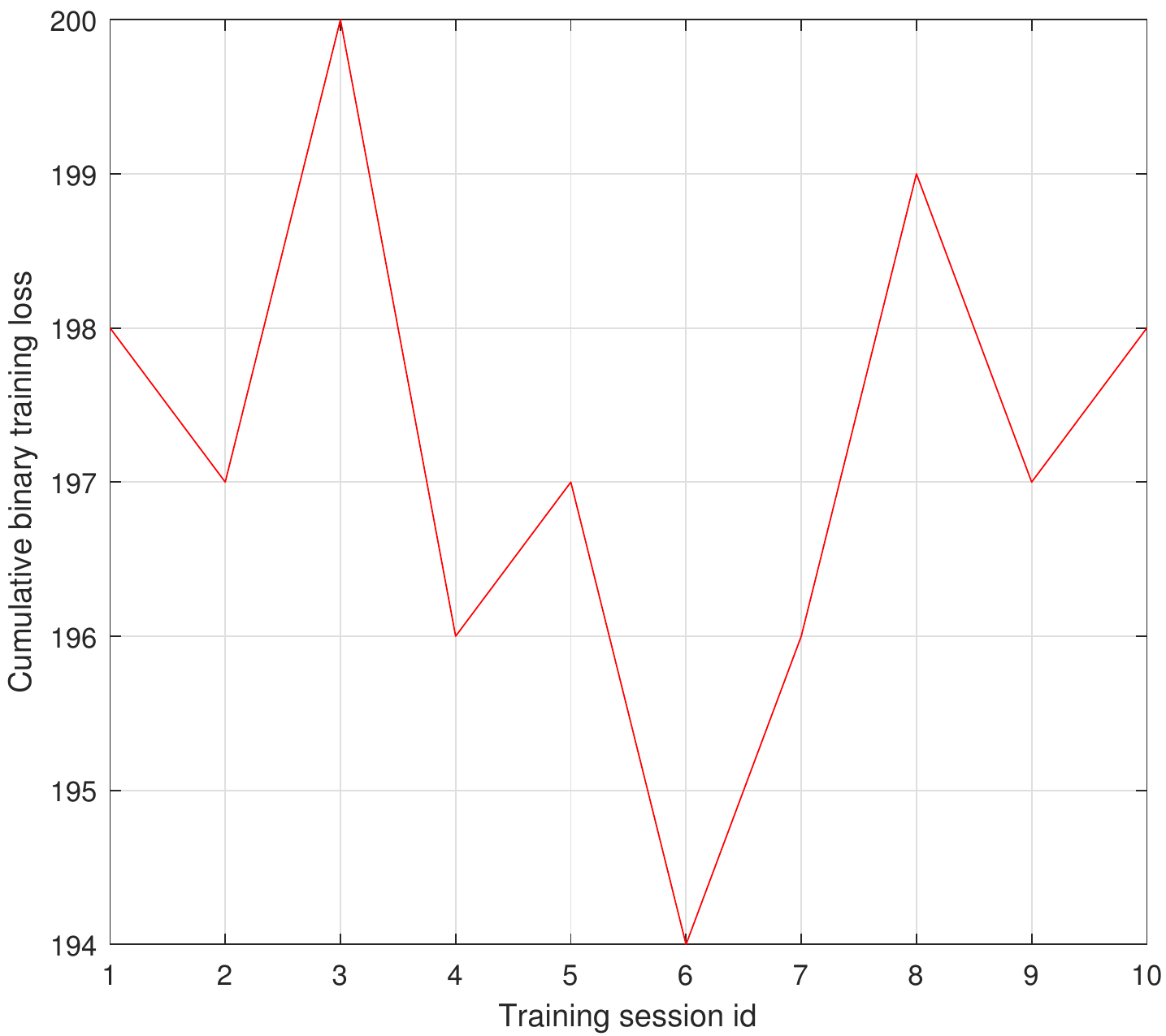}
		\end{minipage}
		\label{5c}
	}
	\subfloat[Training \#44701]{
		\begin{minipage}{0.23\textwidth}
			\includegraphics[width=1\textwidth]{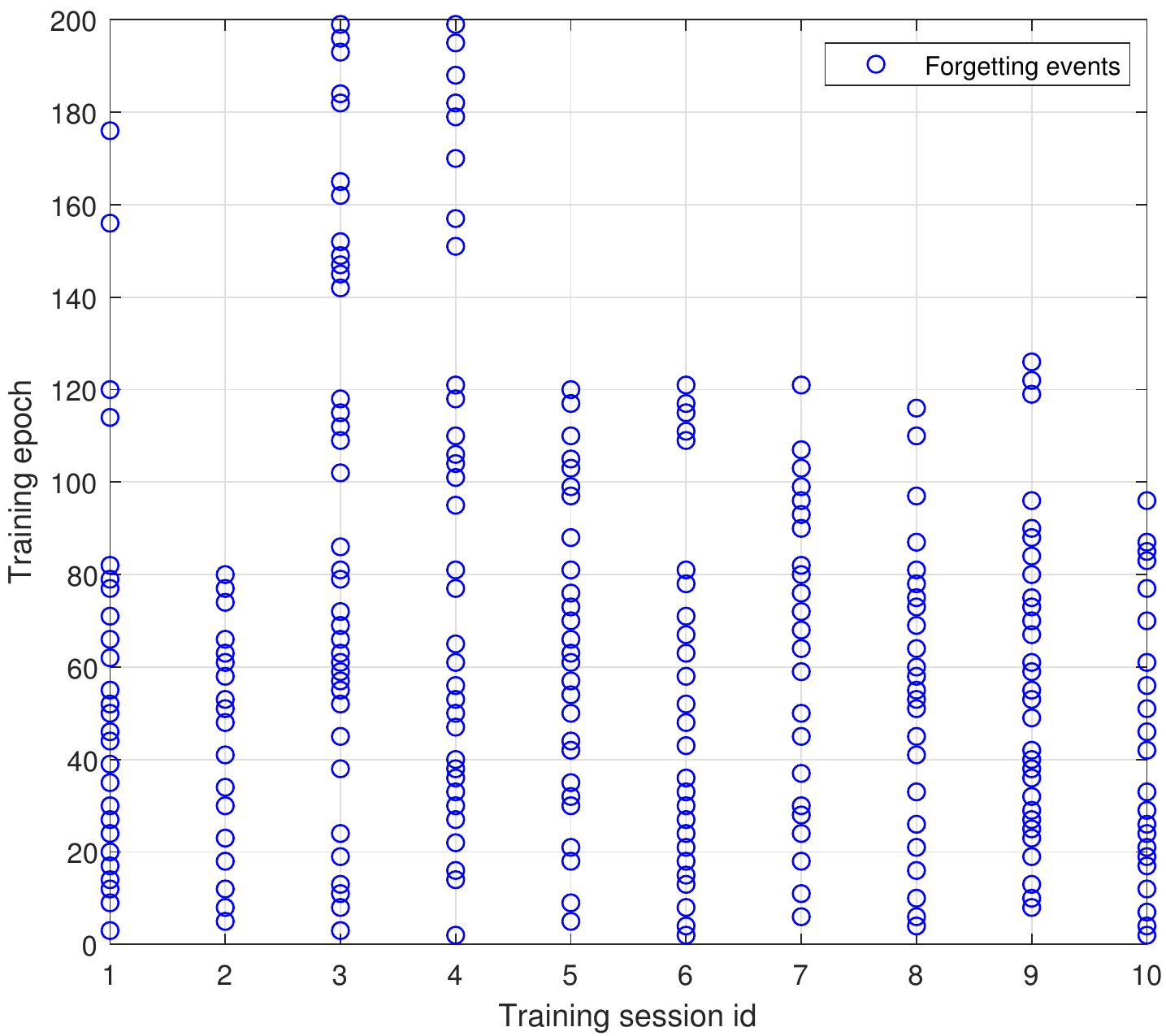} \\
			\includegraphics[width=1\textwidth]{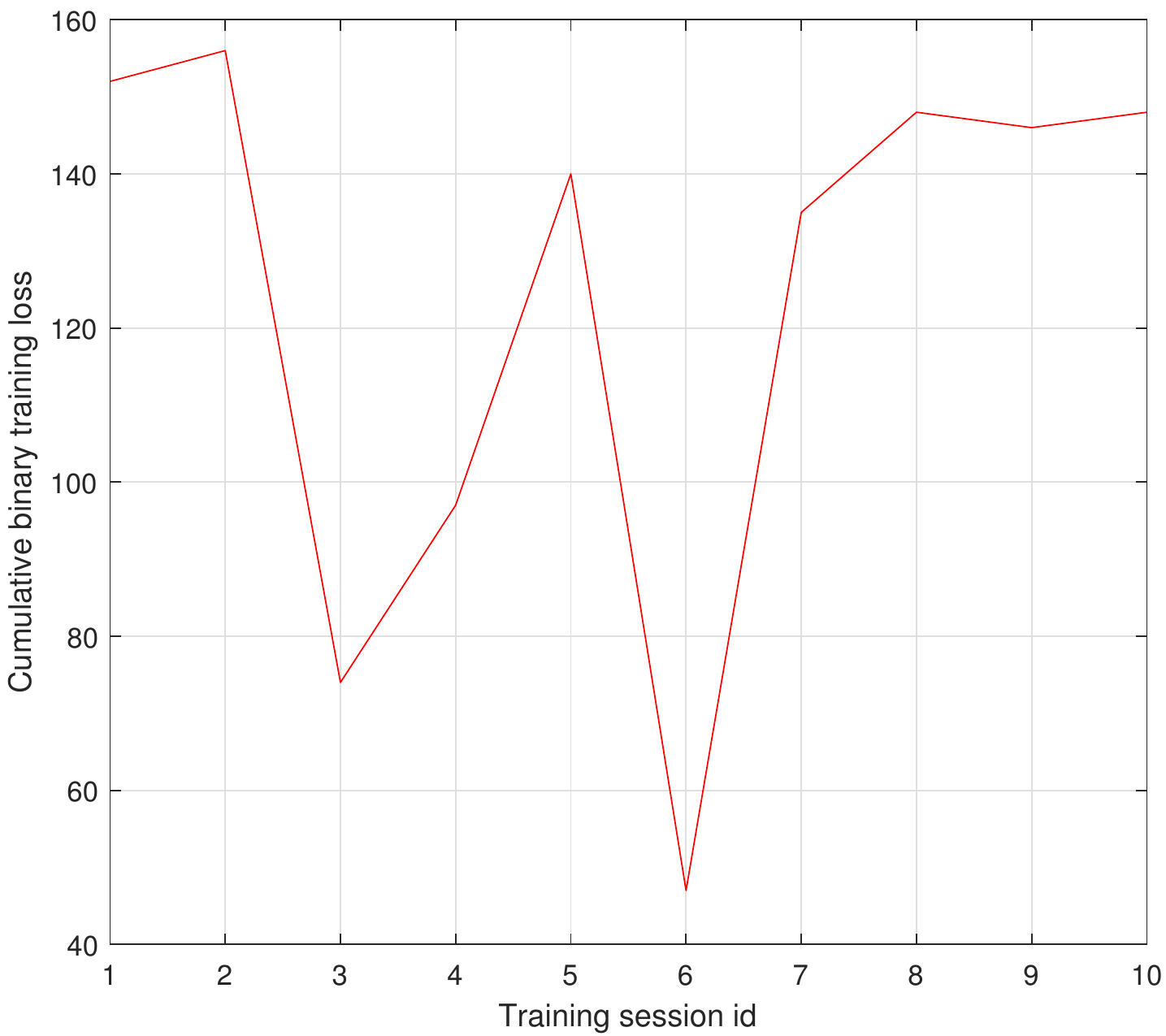}
		\end{minipage}
		\label{5d}
	}
	\caption{The two left columns are dynamic stability of mal-generalizing events and CBGL for testing samples among 10 repetitive training, and the two right columns are dynamic stability of forgetting events and CBTL for training samples. The blue circles in the top row of the figures mark the epoch the mal-generalizing/forgetting event occurs. The number after \# in the caption represents the id of the sample in the Cifar-10 dataset.}
	\label{5}
\end{figure*}

As shown in the first row of Figure \ref{5}, the epoch and the number of the mal-generalizing events' occurrences and the CBGL for the same sample vary in repetitive experiments. For example, over 10 training sessions, the epoch of the mal-generalizing events' occurrences for testing sample \#1 varies between $(0, 90)$, the number varies between $[0, 4]$; and the CBGL varies between $[195, 200]$. The forgetting events and CBTL also have similar properties. This means that  CBTL/CBGL and forgetting/mal-generalizing events should be counted for multiple repetitions of the experiment.

\begin{figure}[!t]
	\centering
	\subfloat[Training samples]{\includegraphics[width=0.4\textwidth]{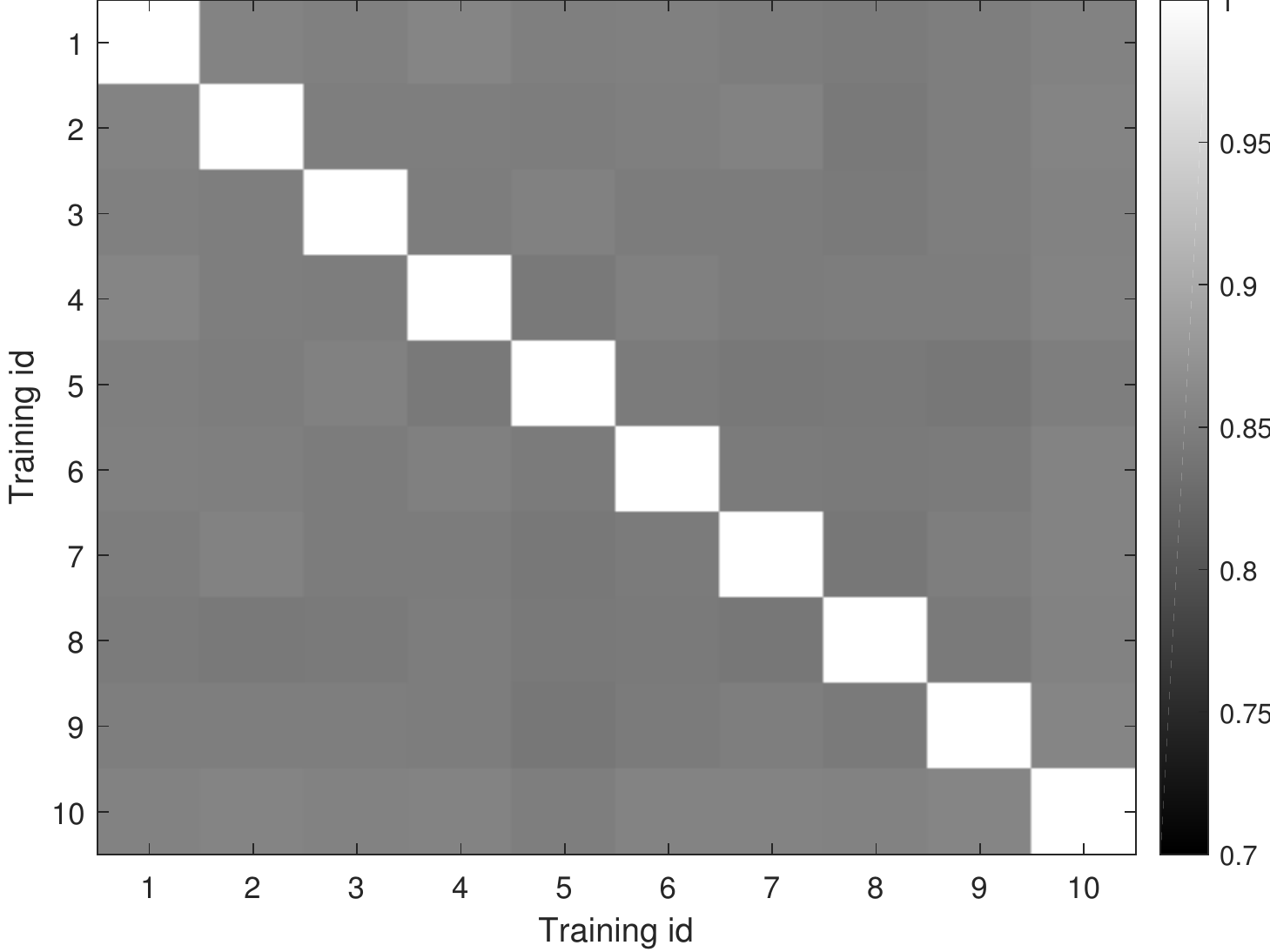}%
		\label{6a}}
	\hfil
	\subfloat[Testing samples]{\includegraphics[width=0.4\textwidth]{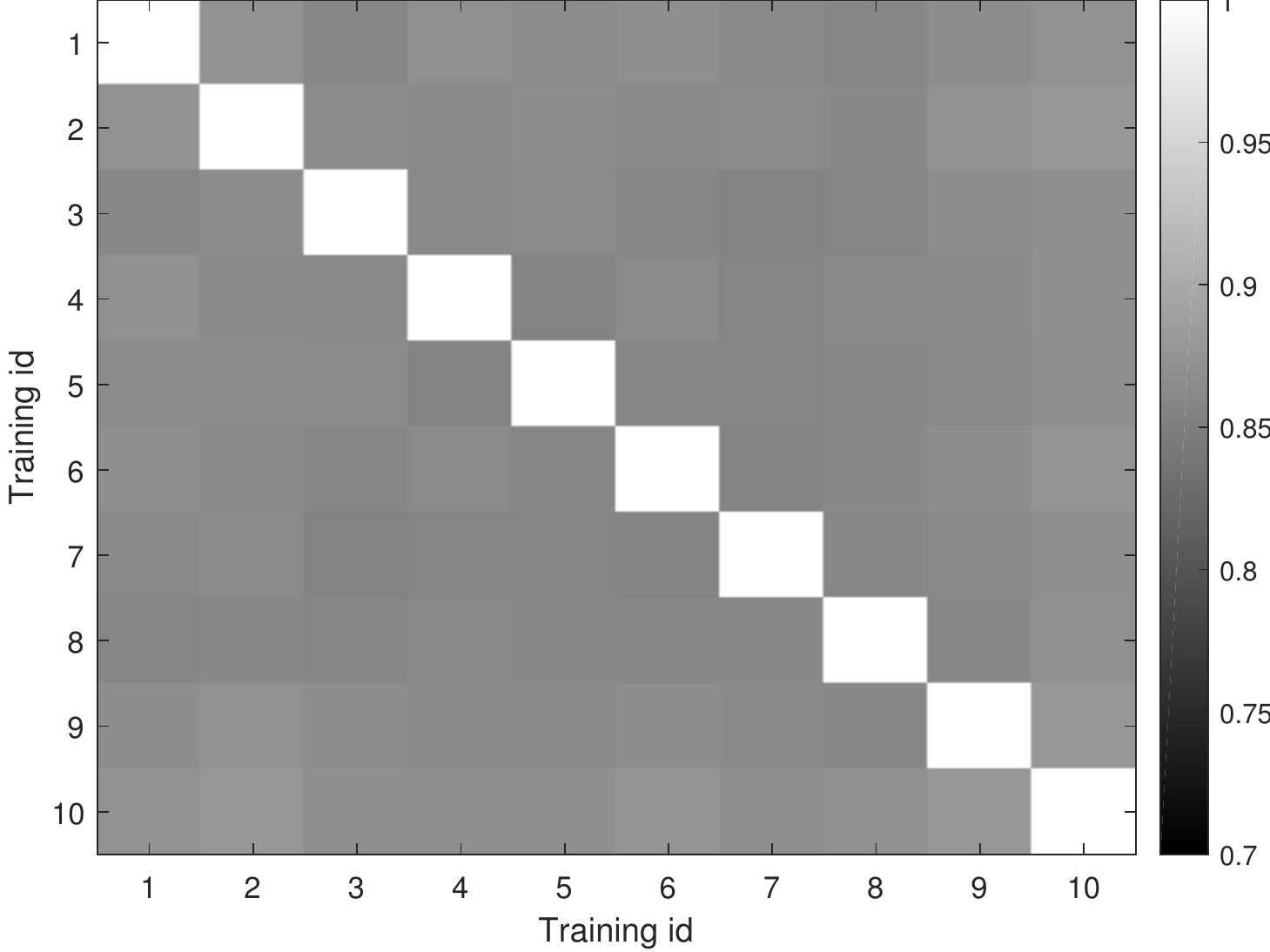}%
		\label{6b}}
	\hfil
	\caption{Correlation of the sample distribution for 10 training sessions. The left is for training samples and the right is for testing samples. The figure shows a $10\times10$ grid, representing the two-by-two correlation of the sample distribution obtained from 10 training sessions. The color shades of each grid represent the values of Pearson correlation coefficient shown in the toolbar on the right side of the figure.}
	\label{6}
\end{figure}

Further, in order to intuitively quantify the effect of randomness from training on the sample distribution representation, we measure the correlation between the results of distributions from different training sessions in this paper. The correlation matrix is obtained based on the correlation between the sample density vectors in each representation as shown in Figure\ref{6}. The sample density vectors is obtained by normalizing the vectors gotten by the density calculation method described in the caption of Figure 2.  

The average Pearson correlation coefficient of 10 training sessions is 0.8634 for training samples and 0.8733 for testing samples. This shows that our representation of the sample distribution has more stable statistical properties for repetitive training sessions.

\subsubsection{The Relationship Between Memorization and Generalization}
Due to the similarities in definitions and the computational process, here we discuss the similarities between mal-generalizing events and forgetting events to understand the relationship between memorization and generalization through statistics of neural network's inference  performance dynamics on training and testing samples.
\begin{figure*}[!t]
	\centering
	\subfloat[Model \#1]{\includegraphics[width=0.5\linewidth]{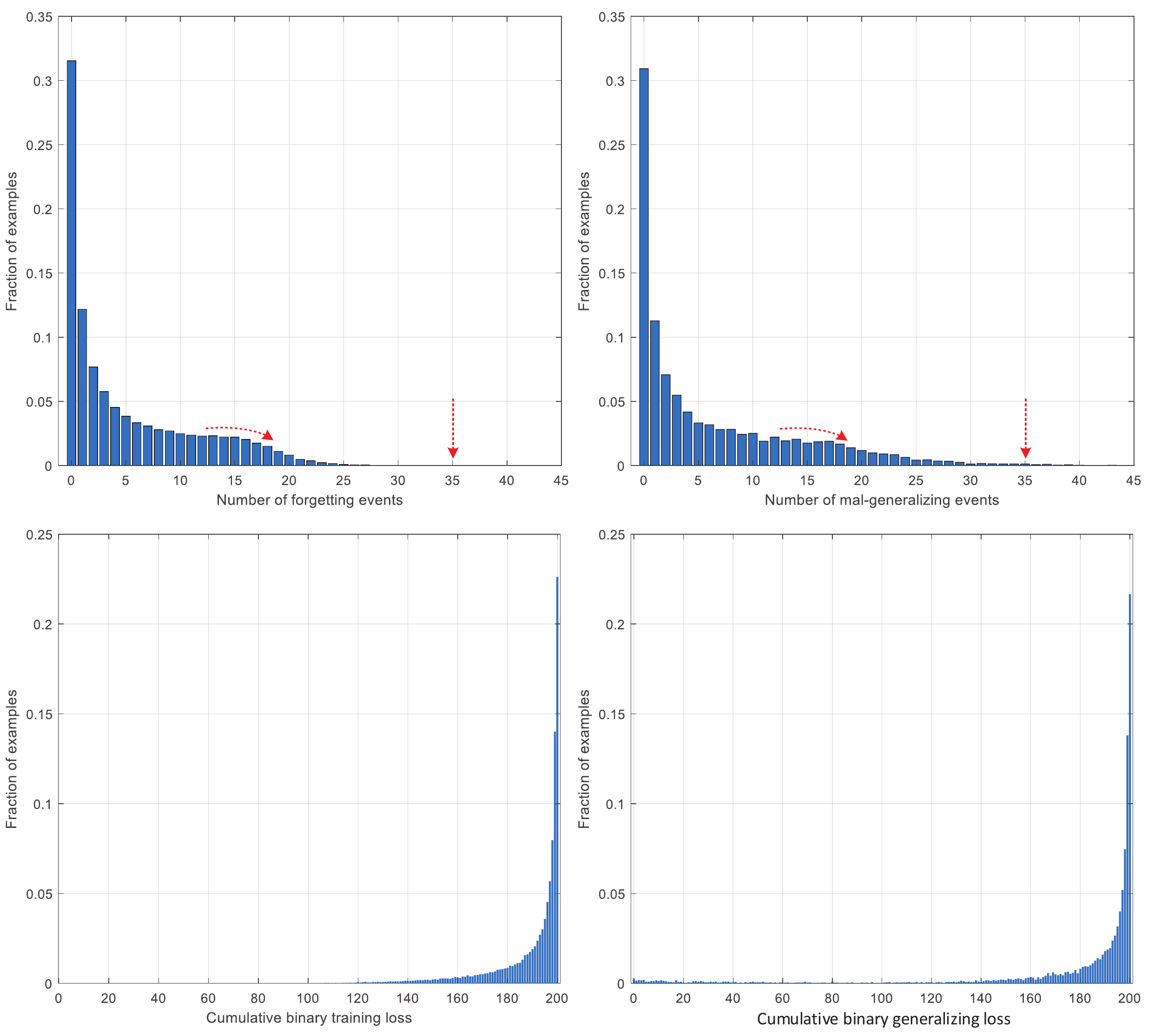}%
		\label{7a}}
	\hfil
	\subfloat[Model \#4]{\includegraphics[width=0.5\linewidth]{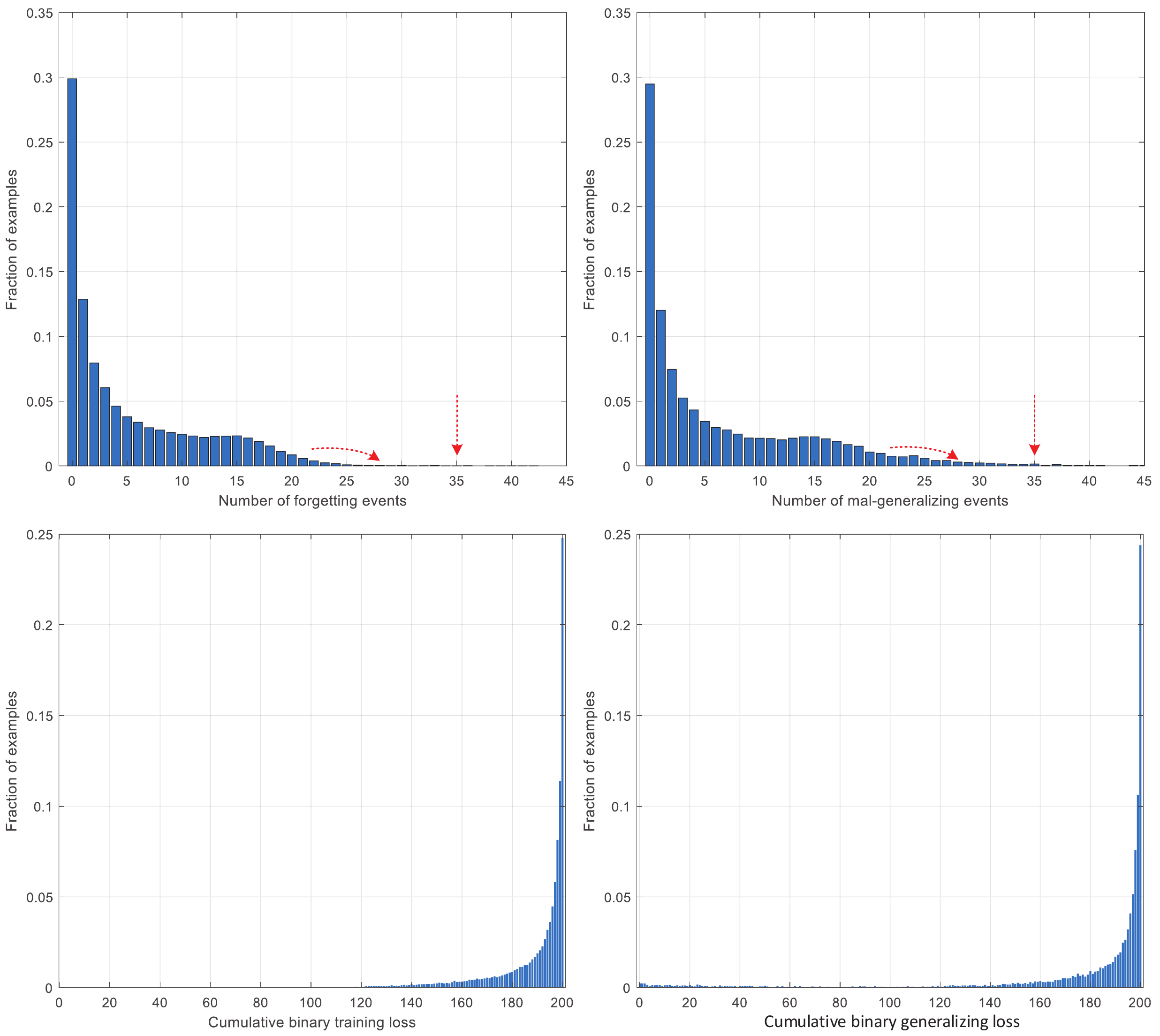}%
		\label{7b}}
	\hfil
	\caption{The relationship between forgetting events and mal-generalizing events. $\#1$ indicates the model obtained from the 1st repetitive training session of the ResNet-110 architecture and $\#4$ is the same. The red arrow indicates the variability of the distribution.}
	\label{7}
\end{figure*}

As shown in Figure \ref{7}, the histogram distribution of forgetting events is similar to that of mal-generalizing events for model $\#1$, as well as for model $\#4$, with Pearson correlation coefficients of 0.9985 and 0.9984, respectively. Likewise, the histogram distribution of CBTL is similar to that of CBGL in the experimental statistics of each model. According to the record\citeup{cifar10} for producing Cifar-10 dataset, the training and testing sets are independently and identically distributed. This means that ideally, the similarity between the distribution of forgetting events and mal-generalizing events of the model can be used to measure the similarity between the training and testing sets. In addition, the red arrows indicate the distribution variability, which to some extent reflects the difference between model learning and generalizing, i.e., generalization error.

\begin{figure}
	\centering
	\includegraphics[width=1\linewidth]{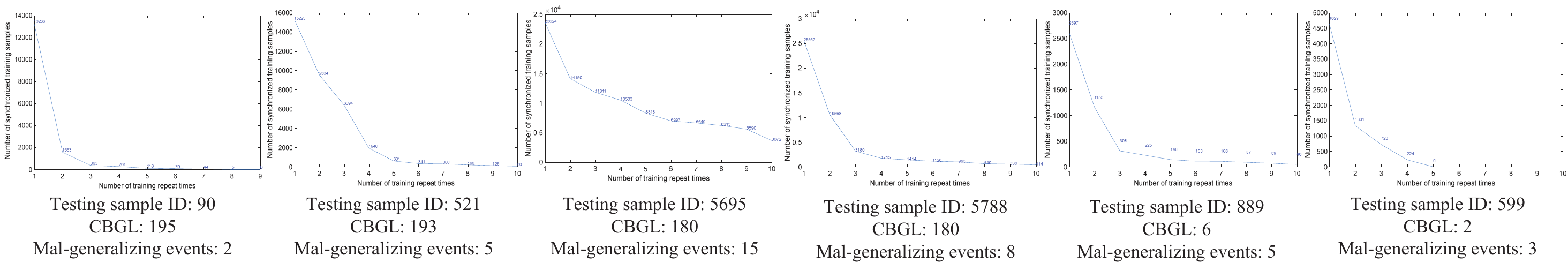}
	\caption{The synchronization between mal-generalizing events and forgetting events}
	\label{4-8}
\end{figure}

Further, we investigate the synchronization between mal-generalizing events and forgetting events, where the synchronization of the particular testing sample with a training sample means that the epoch when its mal-generalizing event occurs during the training process corresponds exactly to the epoch when the forgetting event of that training sample occurs. Testing samples that generalize successfully or unsuccessfully in all training epochs do not have synchronization with any training sample because there are no mal-generalizing events happening. Thus, we are first concerned with the synchronization in different generalization cases. As shown in Figure \ref{4-8}, the number of synchronized training samples varies with the number of mal-generalizing events for the testing samples. Specifically, the number of synchronized training samples is higher in the case of a high number of mal-generalizing events. For example, for sample 90 (with 2 mal-generalizing events ), the number of synchronized training samples is 13266; while sample 5695 (with 15 mal-generalizing events) has about twice as many synchronized training samples during a single training.

\begin{figure}
	\centering
	\includegraphics[width=1\linewidth]{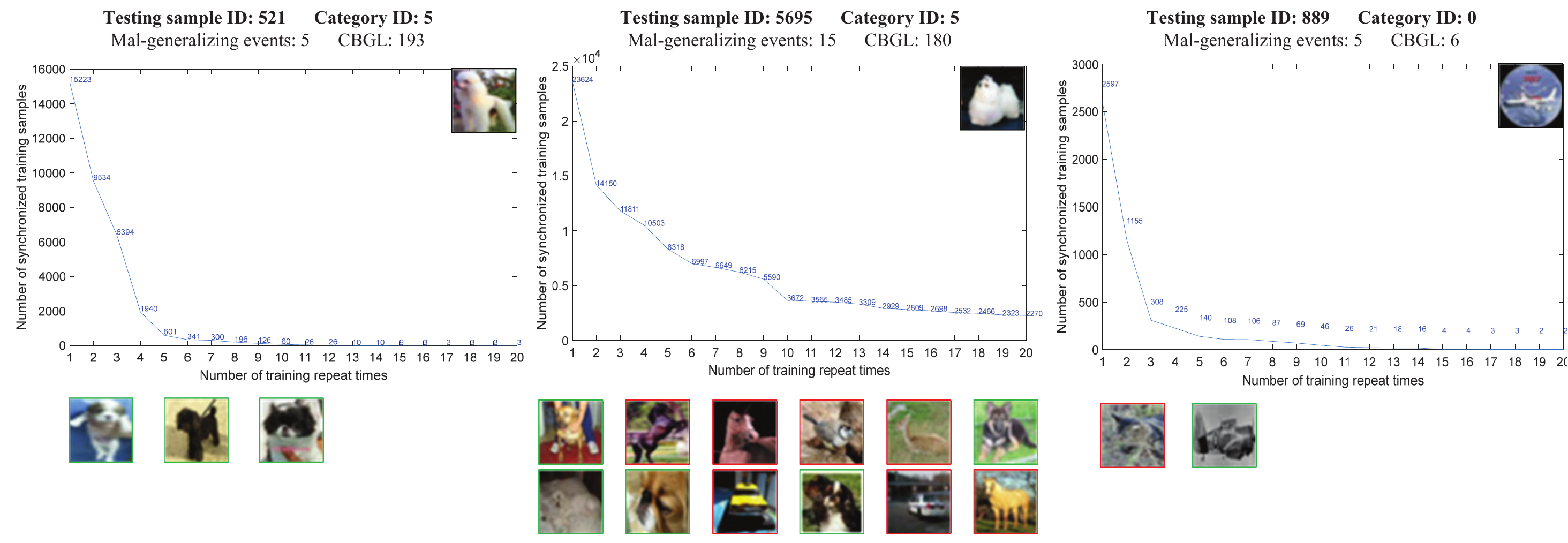}
	\caption{Long-term synchronization between mal-generalizing events and forgetting events. Green and red boxes indicate training samples that belong to the same and different categories as the testing sample, respectively.}
	\label{4-9}
\end{figure}

Considering the aforementioned randomness in the training and generalization process, it is also recorded in Figure \ref{4-8} that the number of synchronized training samples for a particular testing sample decreases significantly, or even drops to 0, when extended to 10 trainings. This indicates that most of the forgetting events of synchronized training samples just happen to occur at the same epoch as the mal-generalizing events of the target testing sample. Although this synchronization occurs more than hundreds of times. This confirms that the neural network does not depend on specific training samples for the extraction and generalization of patterns embedded in the training data. Then does the generalization of specific irregular samples depend on memorizing certain training samples? We extend to 20 trainings for samples 521, 5695, and 889 to observe the distribution of their synchronized training samples, as shown in Figure \ref{4-9}.

To further verify the effect of these synchronized training samples on the generalization of the corresponding testing samples, we train a binary classifier using SVM to determine whether a particular testing sample belongs to a certain class. Its training data input is 4096-dimensional features extracted from the training samples using the VGG-19 network pre-trained by ImageNet. The experimental setup is divided into two types: $1.$ random $N$ training samples containing 46 synchronized samples ($50\%$ each of training samples belonging to the same and different categories as the testing sample, same later); $2.$ random $N$ training samples not containing these synchronized samples. The results are shown in Table \ref{table4-1}, where the classifier trained with synchronized samples can better generalize its corresponding testing samples for sample 889 with low-regularity. For example, when the number of synchronized samples accounts for $46\%$ of the training samples, the classification accuracy of the classifier trained with synchronized samples for sample 889 is about 14 percentage points higher than that of the classifier trained without synchronized samples. The lead is still about 3 percentage points when the proportion of synchronized samples is reduced to $10\%$. We conjecture that these synchronized training samples tend to play the role of "support vectors". That is, stochastic gradient descent tends to converge implicitly to the solution that maximizes the differentiation of the dataset \citeup{137}, and the synchronized training samples are more relevant to the learning of the corresponding testing samples' classification boundaries than the other samples.

 \begin{table}
	\caption{An exploration of the supportiveness of synchronized training samples to testing samples} 
	\label{table4-1}
	\resizebox{\textwidth}{!}
	{
	\begin{tabular}{lllll}
		\hline\noalign{\smallskip}
		\multirow{2}*{Sample ID} & \multicolumn{2}{c}{N=100, 46 synchronized training samples} &\multicolumn{2}{c}{N=460, 46 synchronized training samples}  \\
		\cline{2-5}
		& w/ synchronized samples &w/o synchronized samples & w/ synchronized samples &w/o synchronized samples\\
		\noalign{\smallskip}\hline\noalign{\smallskip}
		521  & 0.9585 &0.9737 &0.9950 &0.9768\\
		889 & 0.3528 &0.2147 &0.1561 &0.1238 \\
		5695& 0.7448 &0.7496 &0.6470 &0.5107\\
		\noalign{\smallskip}\hline
	\end{tabular}
}
\end{table}

\subsubsection{Robustness}
The generalization of a neural network varies across different optimizers and architectures, so the statistics of the training depend on the given network architecture trained with the given optimizer. It is thus important to explore the effect of these control variables on these four event statistics, and we further explore the robustness of these statistics.

\begin{figure*}[!t]
	\centering
	\subfloat[ResNet-20]{
		\begin{minipage}{0.23\textwidth}
			\includegraphics[width=1\textwidth]{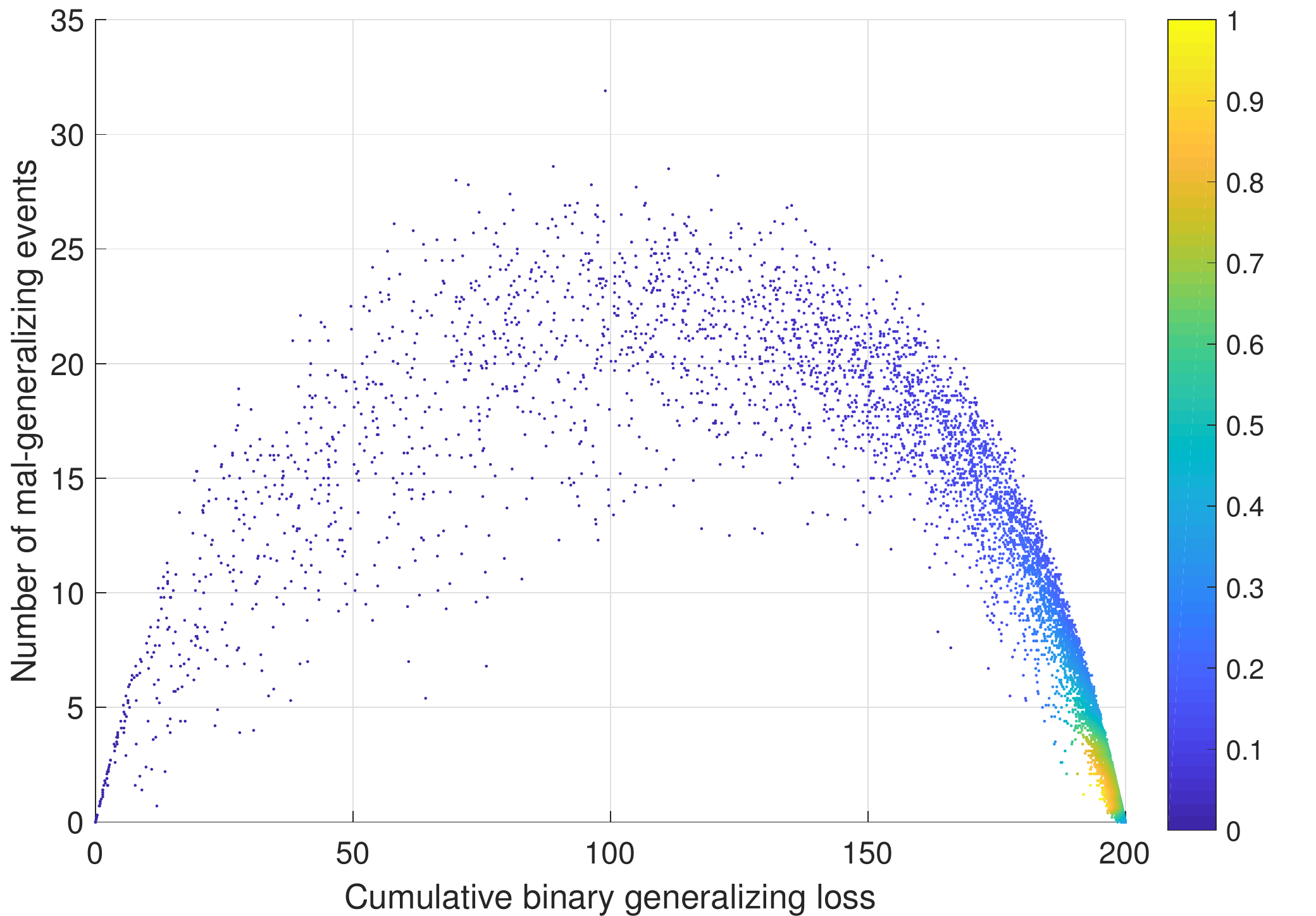} \\
			\includegraphics[width=1\textwidth]{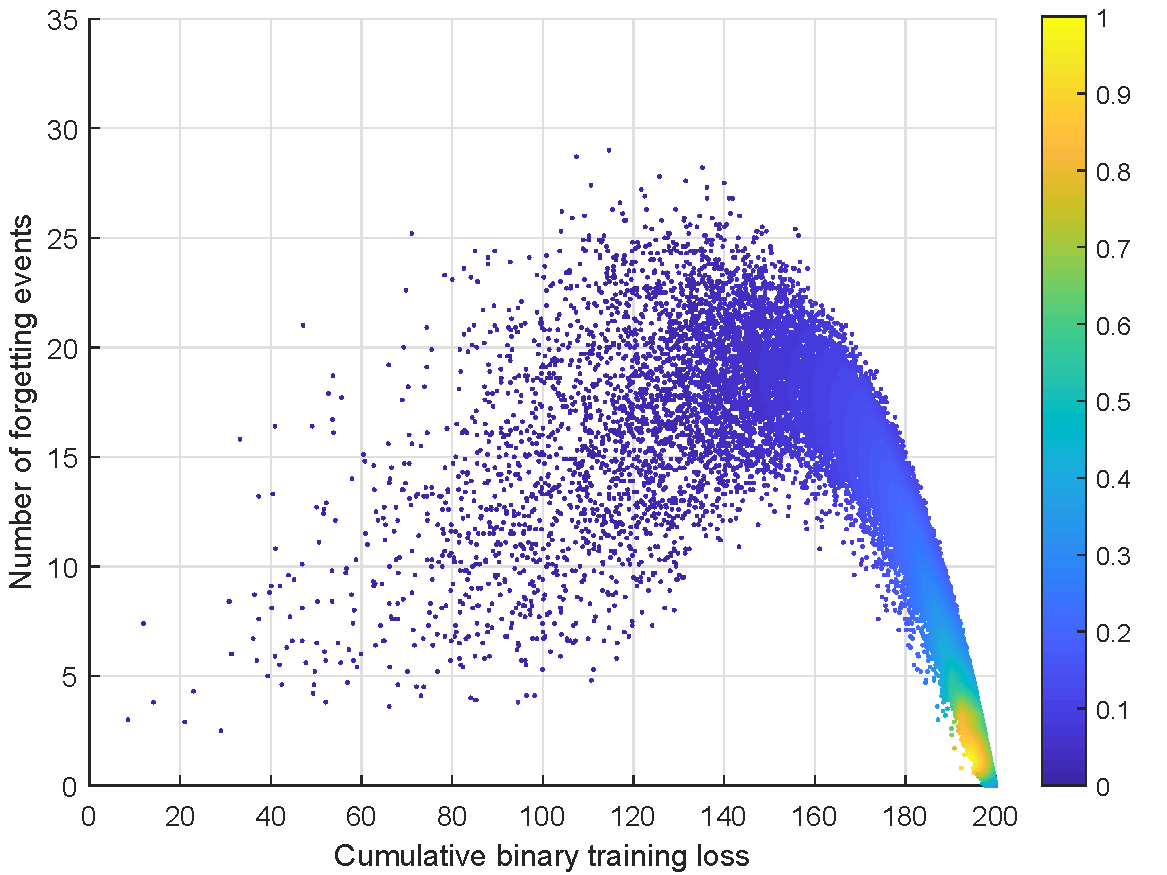}
		\end{minipage}
		\label{8a}
	}
	\subfloat[ResNet-32]{
		\begin{minipage}{0.23\textwidth}
			\includegraphics[width=1\textwidth]{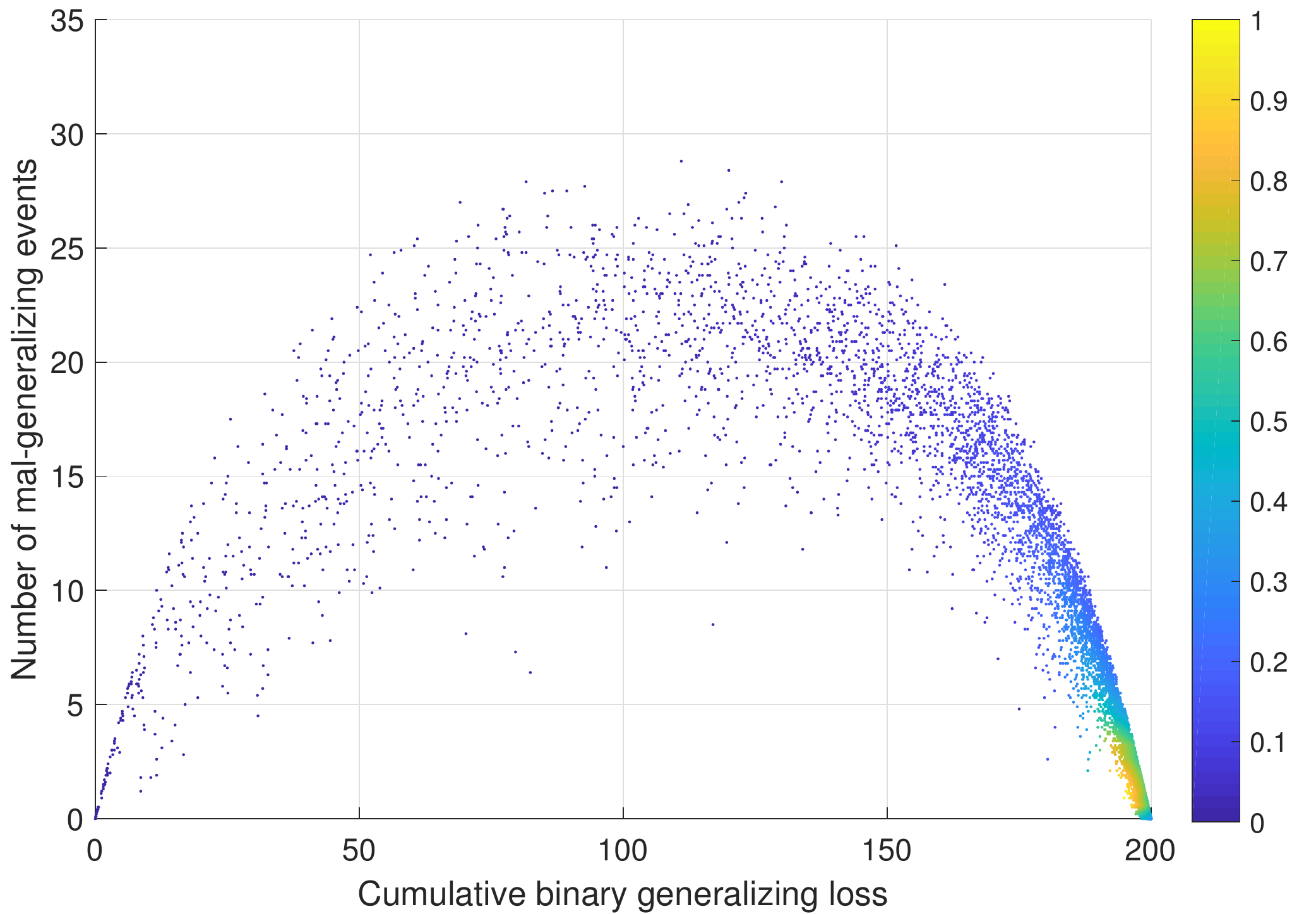} \\
			\includegraphics[width=1\textwidth]{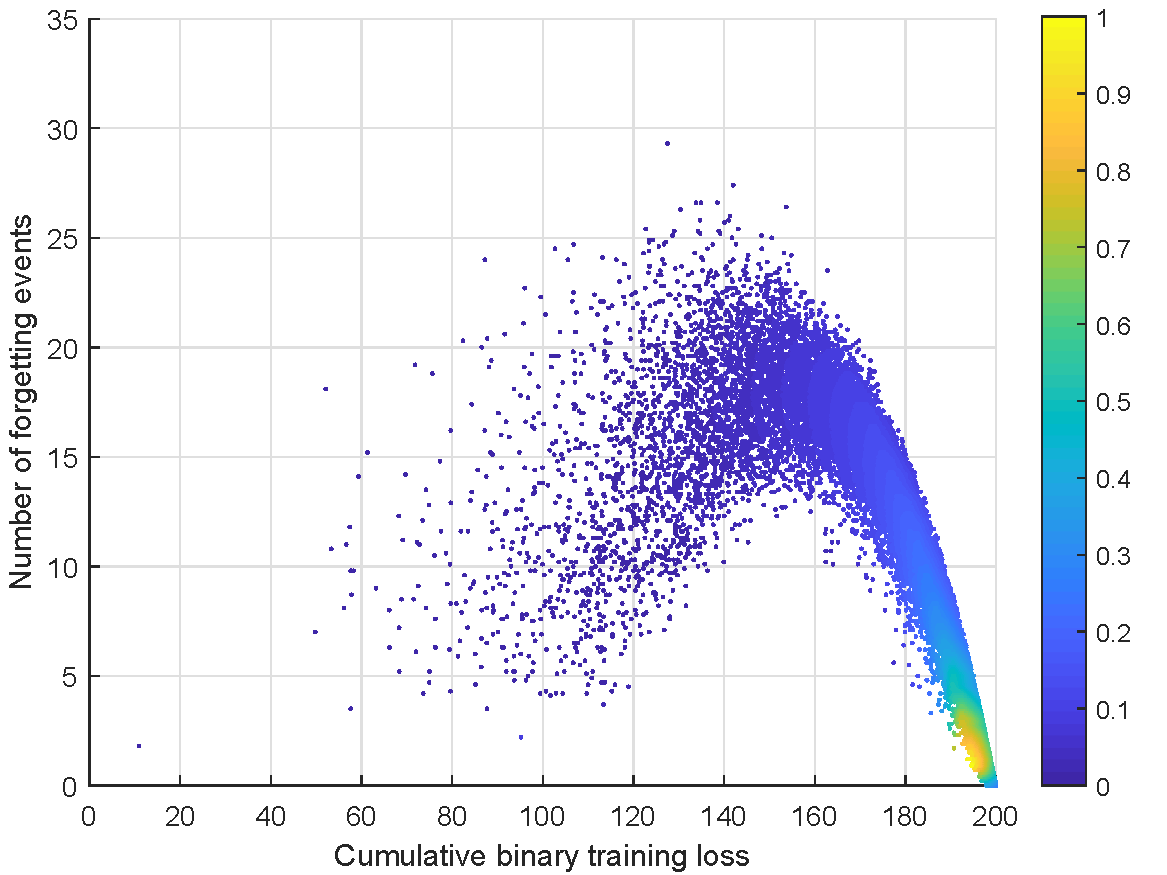}
		\end{minipage}
		\label{8b}
	}
	\subfloat[ResNet-110]{
		\begin{minipage}{0.23\textwidth}
			\includegraphics[width=1\textwidth]{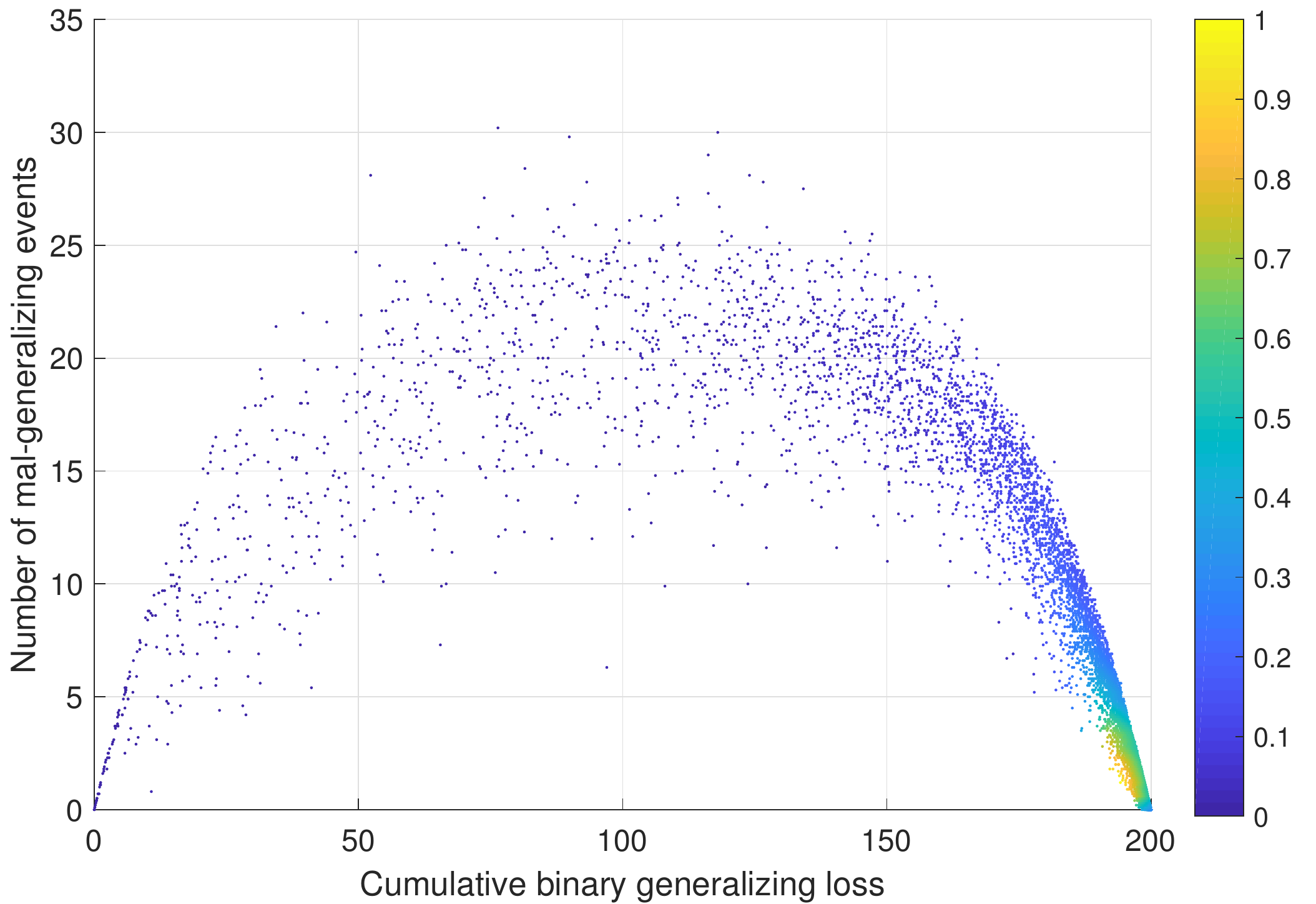} \\
			\includegraphics[width=1\textwidth]{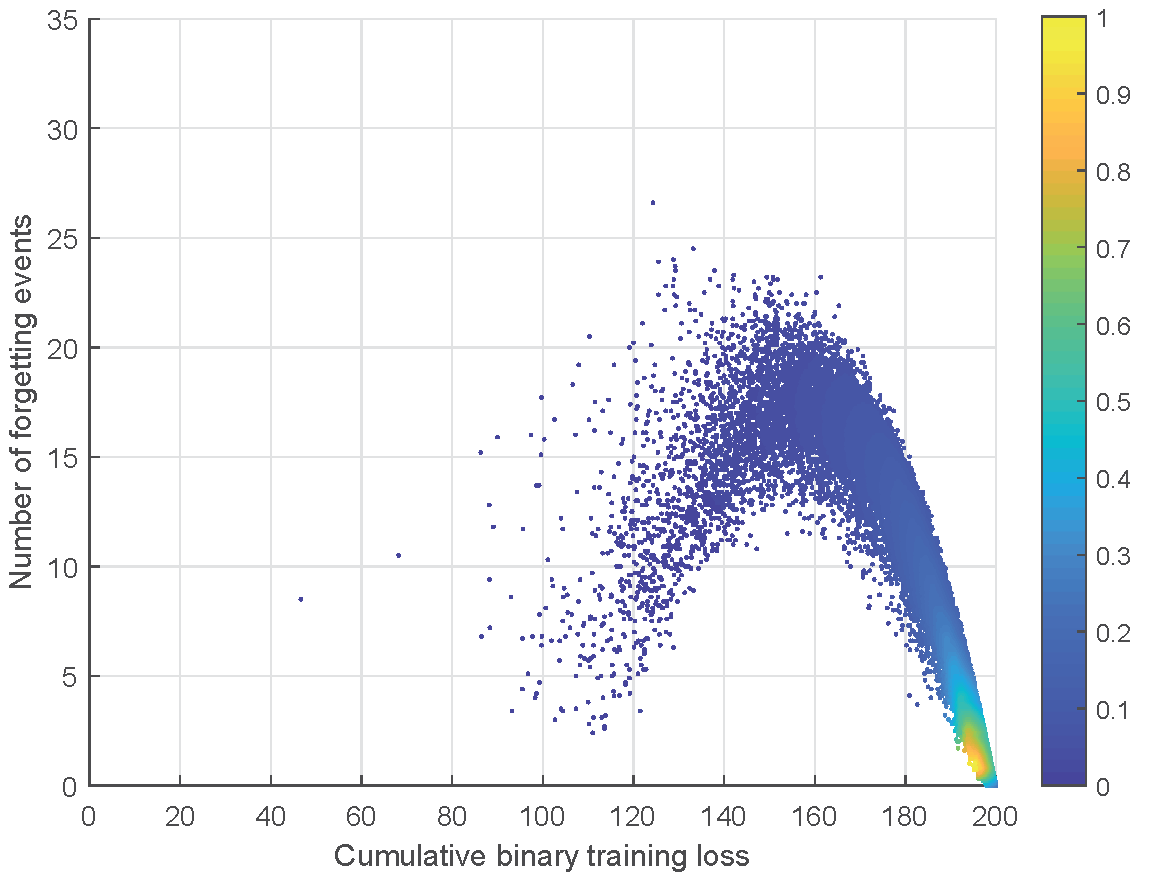}
		\end{minipage}
		\label{8c}
	}
	\subfloat[DenseNet]{
		\begin{minipage}{0.23\textwidth}
			\includegraphics[width=1\textwidth]{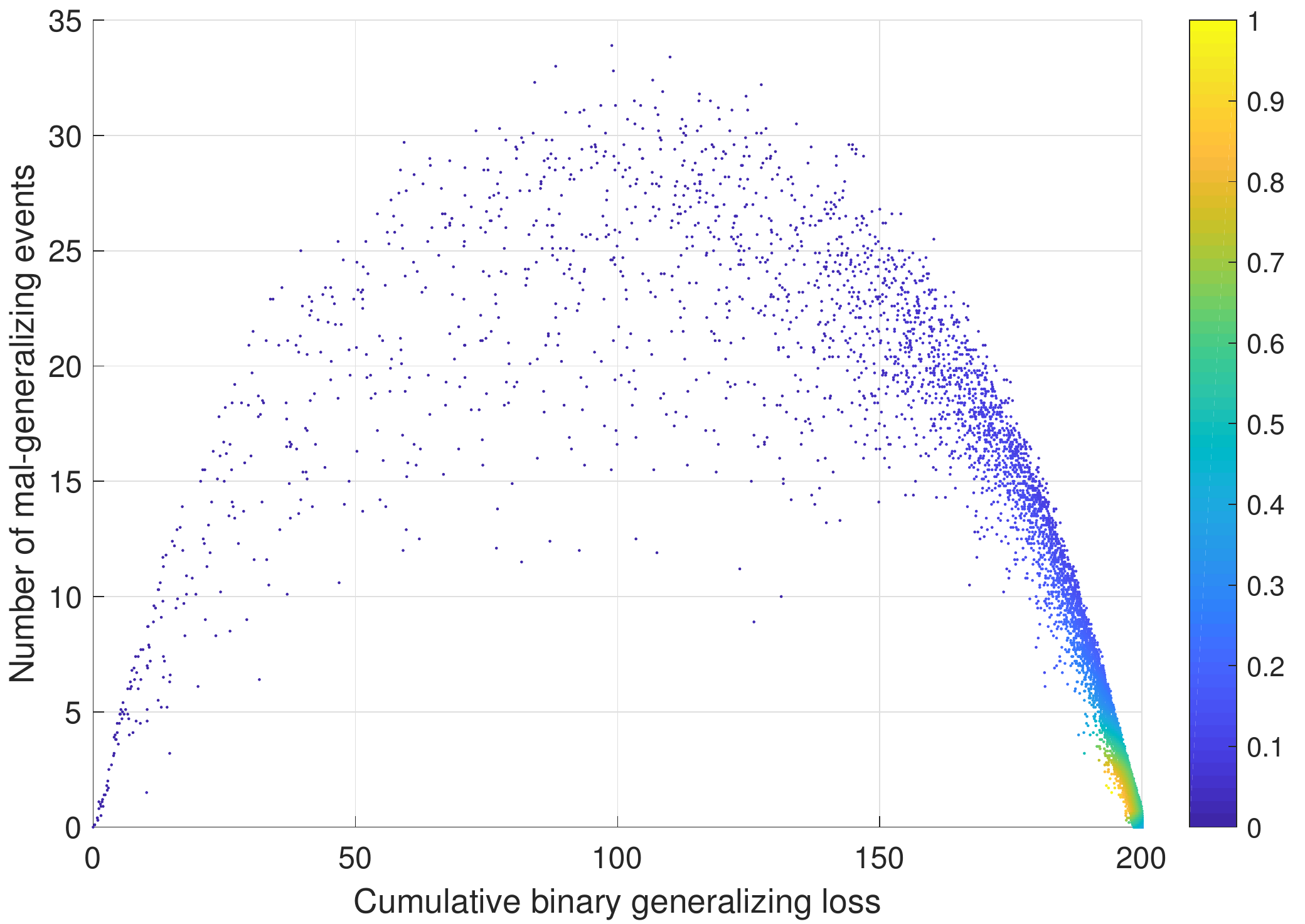} \\
			\includegraphics[width=1\textwidth]{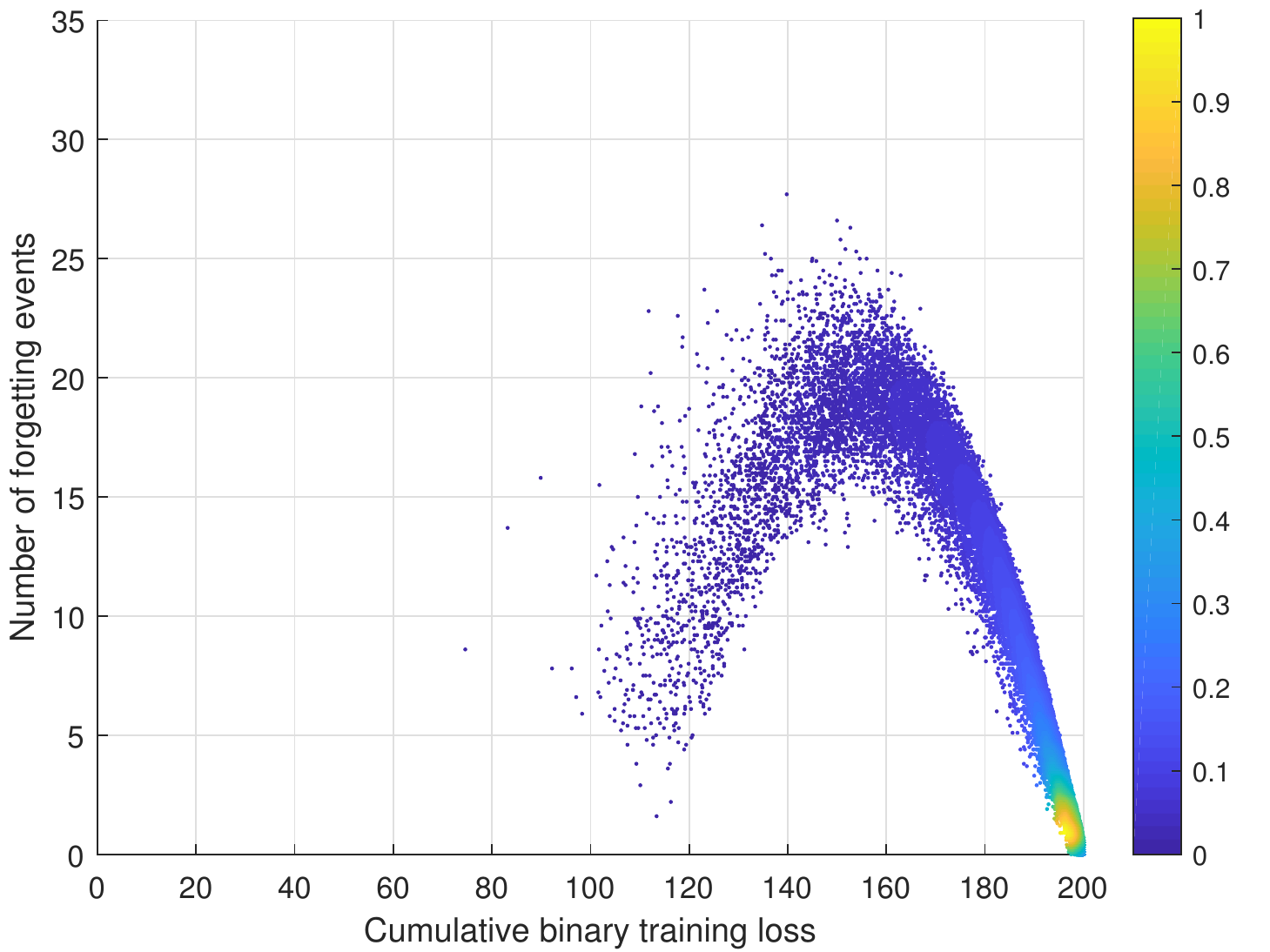}
		\end{minipage}
		\label{8d}
	}	
	\caption{Bi-dimensional sample representations of different network architectures. The top row shows the testing sample representations and the bottom row shows the training sample representations.}
	\label{8}
\end{figure*}

\paragraph{Different Architectures}: Our statistics need to be collected during training, and if the network has more layers or a complex structure, the time cost of statistics will be more expensive. Therefore, we explore how this distribution representation is affected by different network architectures. We hope to see whether the statistic is robust to different network architectures in order to explore the possibility of reliably representing the sample distribution with simpler architectures with lower time cost. We repeatedly trained each of the four network architectures, ResNet-20, ResNet-32, ResNet-110, and DenseNet, 10 times and took the average of statistics for analysis. As shown in Figure \ref{8}, the sample distribution does not change significantly whether the number of layers is reduced from 110 to 32, 20, or to the more densely connected DenseNet. This conclusion is confirmed quantitatively by the correlation calculations of the distributions and statistics. As shown in the table \ref{table1}, the Pearson correlation coefficients are all above 0.88, which are very strong correlations. This indicates that our proposed sample distribution representation is robust to the network architecture and can be migrated between different architectures. Since ResNet-32 takes only 80 minutes while DenseNet takes up to 1367 minutes with the same computational power (single GEFORCE RTX 2080Ti GPU). This opens up the prospect of computing sample distribution representation statistics in simpler architectures. Thus, the similarity of 2D sample distribution representations across architectures allows them to be computed quickly by approximating proxy networks to reduce time and computational cost, e.g., estimating sample distribution statistics for DenseNet by ResNet32 can save up to 21.5 hours of training time and reduce the number of parameters by 7.03 megabytes.


\begin{table}
	\caption{The correlation of the distributions and statistics for different network architectures}
	\label{table1}
	\resizebox{\textwidth}{!}
	{
	\begin{tabular}{lllllll}
		\hline\noalign{\smallskip}
		\multirow{3}*{Pearson correlation coefficient} & ResNet-20 & ResNet-20  & ResNet-20  & ResNet-32  & ResNet-32  & ResNet-110  \\ ~ & vs & vs & vs & vs &vs & vs\\
		~ &  ResNet-32 & ResNet-110 & DenseNet & ResNet-110 & DenseNet & DenseNet\\
		\noalign{\smallskip}\hline\noalign{\smallskip}
		CBTL & 0.9986 & 0.9935 & 0.9826 & 0.9977 & 0.9745 & 0.9585\\
		CBGL & 0.9973	& 0.9901 & 0.9841 & 0.9962 & 0.9733	& 0.9533\\
		Forgetting events & 0.9996 & 0.9978	& 0.9953 & 0.9992 & 0.9950 & 0.9925\\
		Mal-generalizing events & 0.9995 & 0.9989 & 0.9894 & 0.9993	& 0.9924 & 0.9914\\		
		Density for training samples & 0.9603 & 0.9480 & 0.9107	& 0.9594 & 0.9057 & 0.9005\\
		Density for testing samples  & 0.9560 & 0.9327 & 0.8846 & 0.9524	& 0.8955 & 0.9025\\
		\noalign{\smallskip}\hline
	\end{tabular}
}
\end{table}

\paragraph{Different Optimizers}: Most deep learning algorithms involve some forms of optimization, generally with the goal of minimizing the loss function for parameter solving. Thus, optimizers play a crucial role in deep learning algorithms, and in this section we explore the effect of different optimizers on the representation of the sample distribution. In addition to the common mini-batch stochastic gradient descent method (SGD), AdaGrad and AdaMax are also chosen as the comparison experimental setup. 

\begin{figure*}[!t]
	\centering
	\subfloat[SGD]{
		\begin{minipage}{0.3\textwidth}
			\includegraphics[width=1\textwidth]{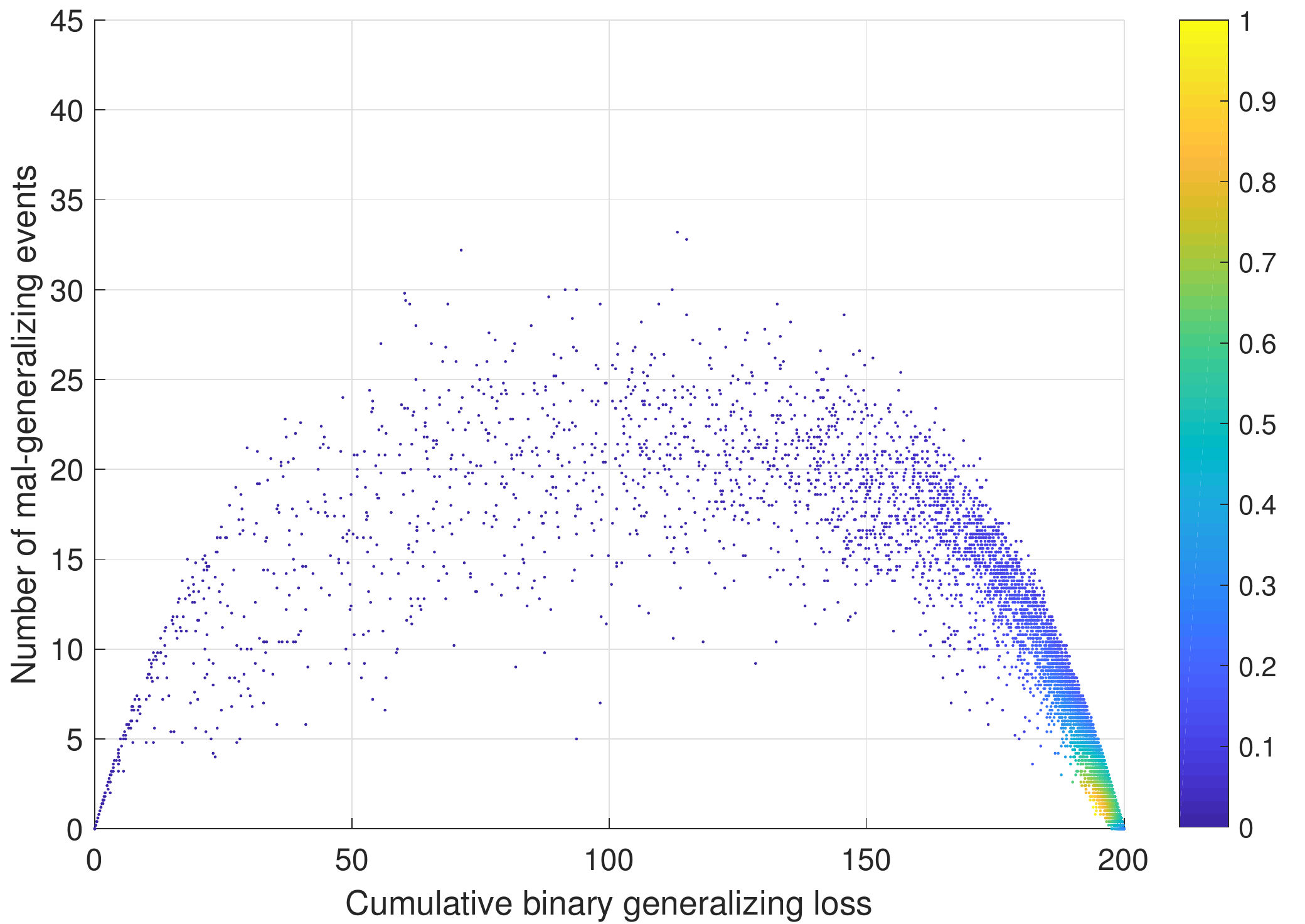} \\
			\includegraphics[width=1\textwidth]{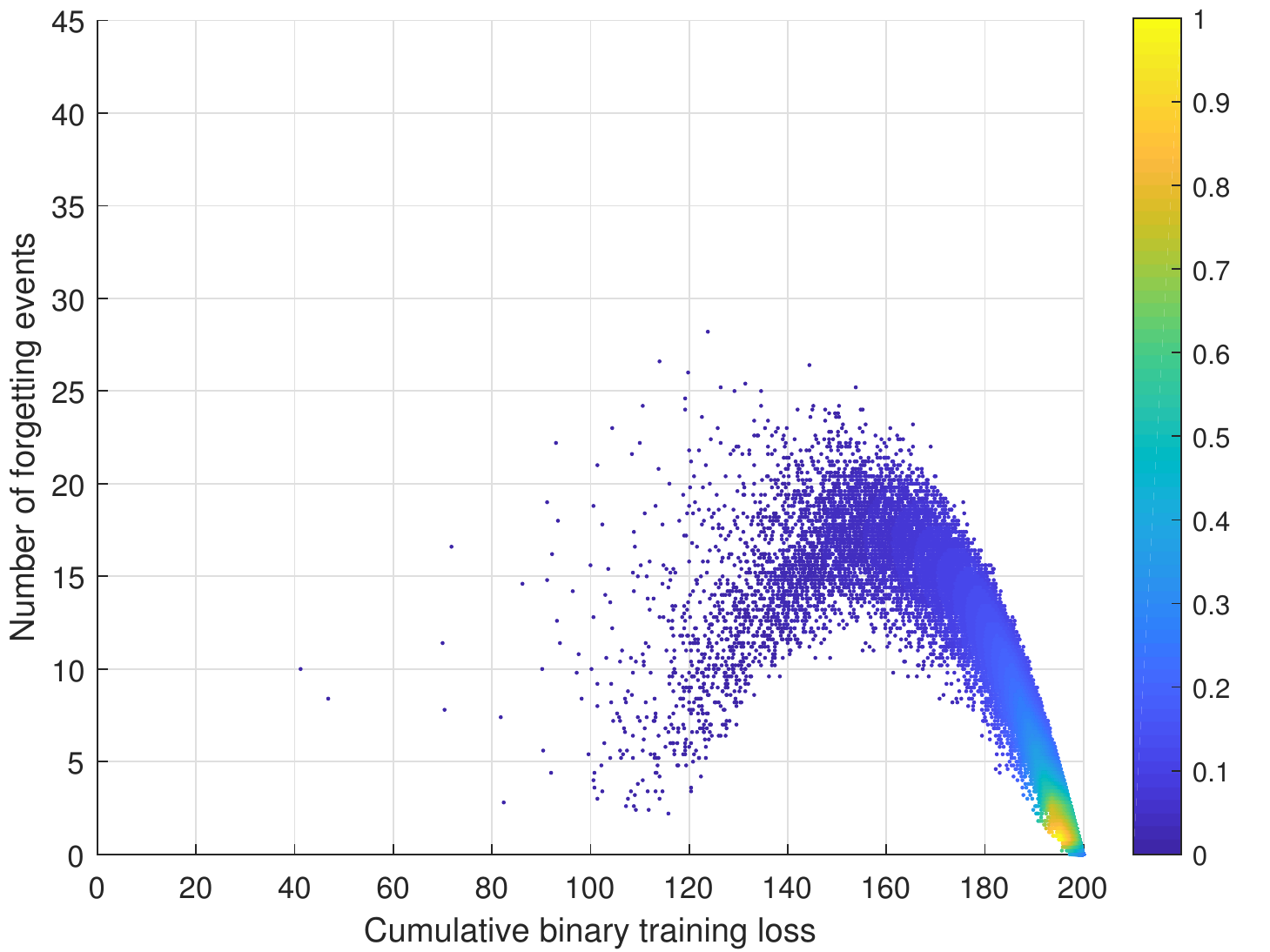}
		\end{minipage}
		\label{9a}
	}
	\subfloat[AdaGrad]{
		\begin{minipage}{0.3\textwidth}
			\includegraphics[width=1\textwidth]{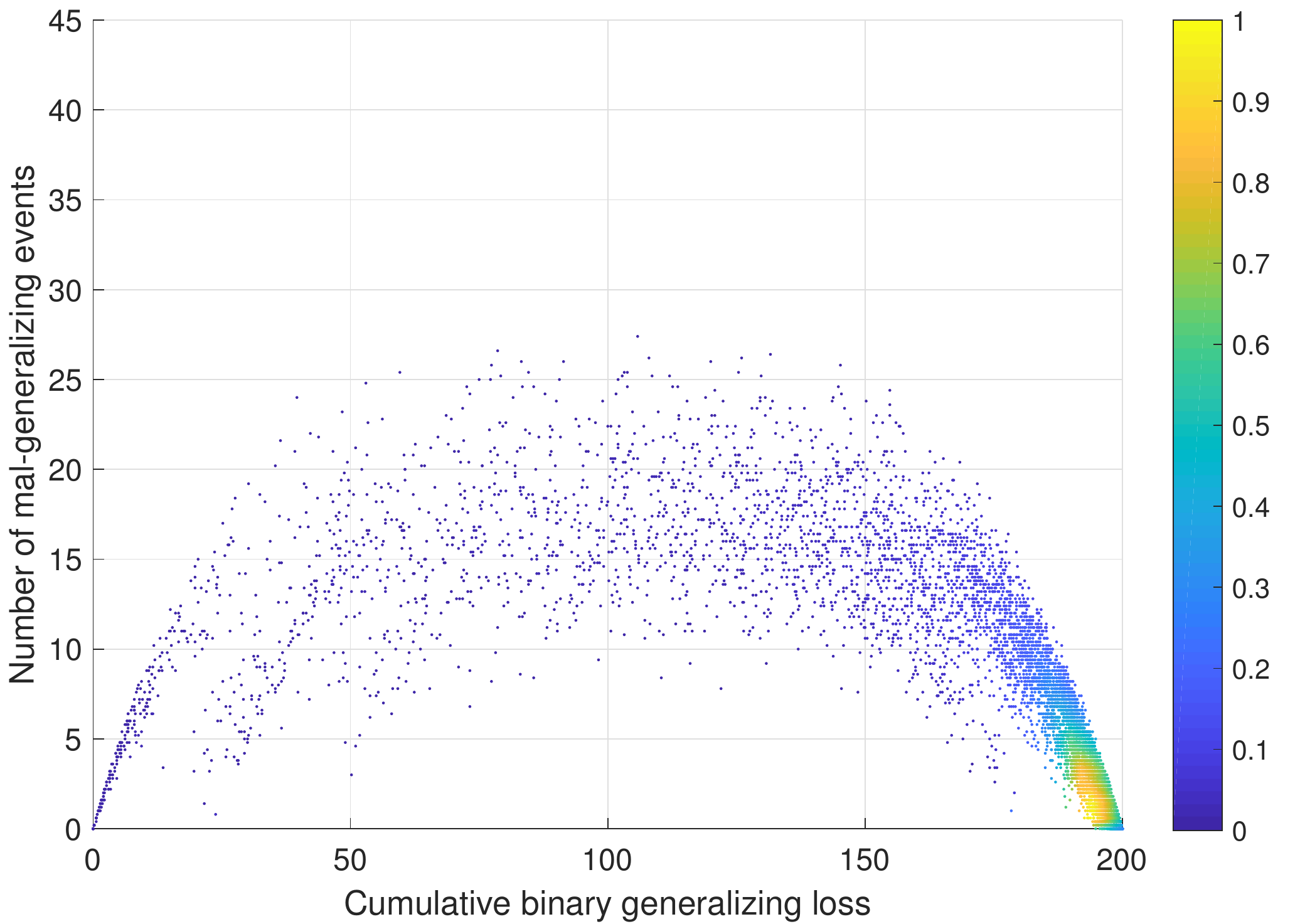} \\
			\includegraphics[width=1\textwidth]{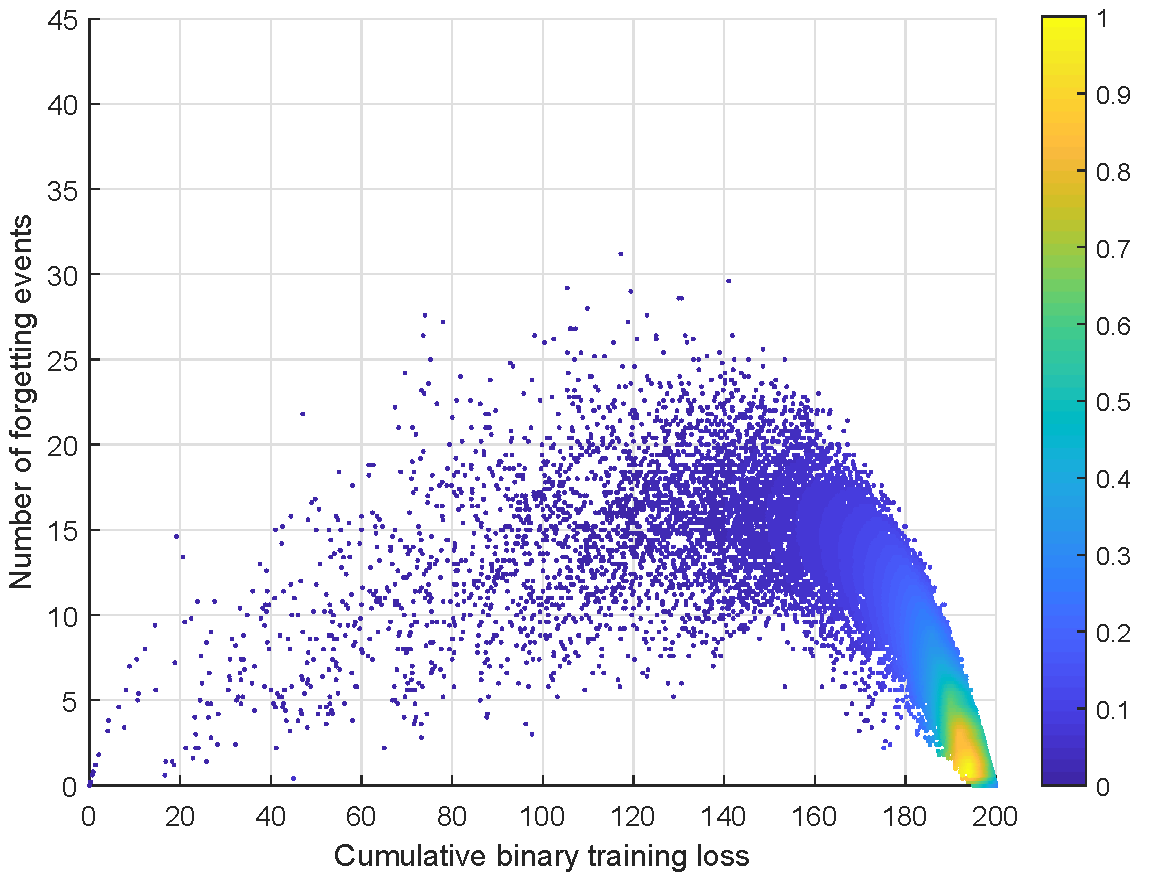}
		\end{minipage}
		\label{9b}
	}
	\subfloat[AdaMax]{
		\begin{minipage}{0.3\textwidth}
			\includegraphics[width=1\textwidth]{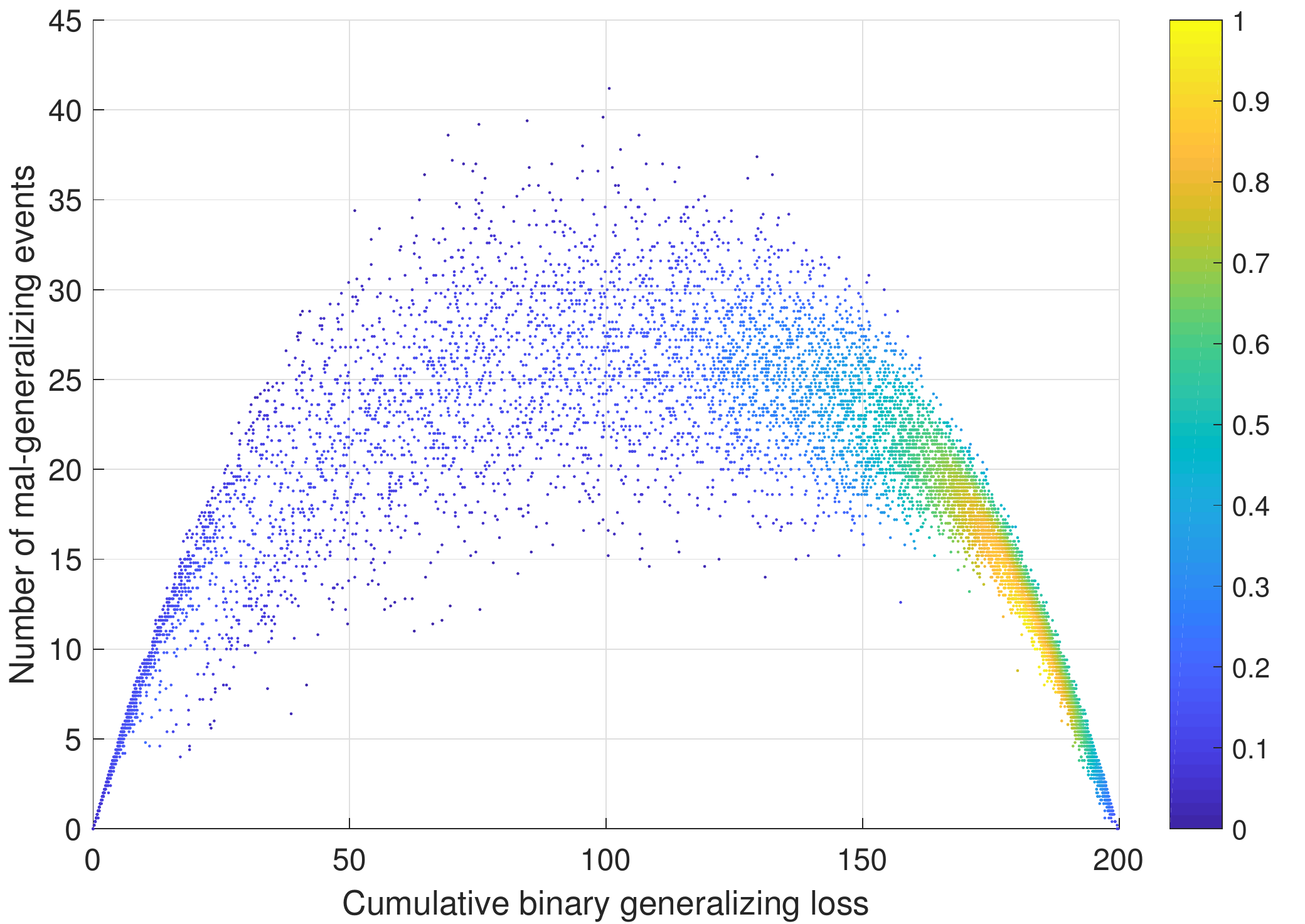} \\
			\includegraphics[width=1\textwidth]{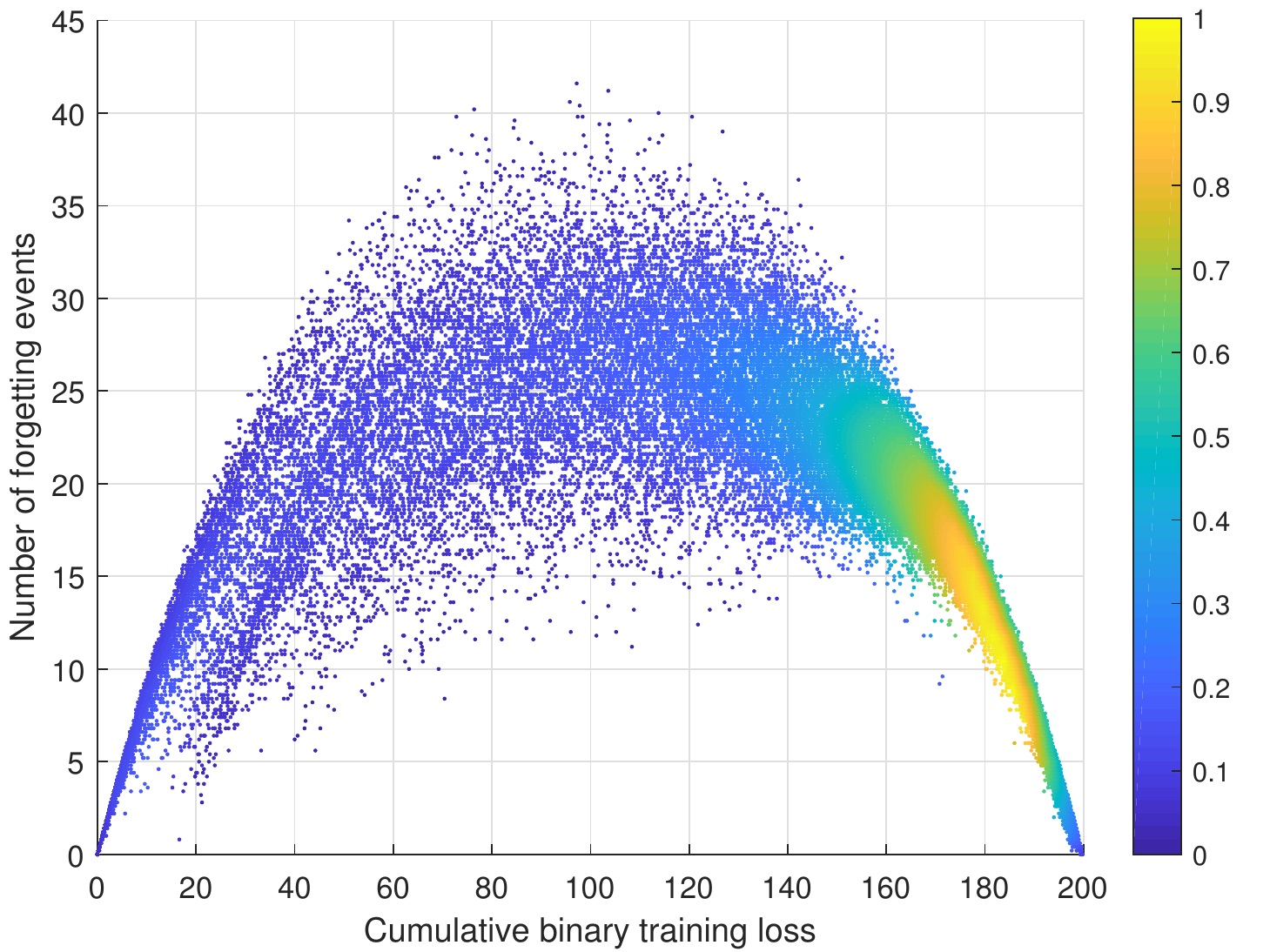}
		\end{minipage}
		\label{9c}
	}
	\caption{Bi-dimensional sample representations of the ResNet-110 trained with different optimizers. The top row is the testing sample representations and the bottom row is the training sample representations. From left to right, each column shows the representation results for SGD, AdaGrad, and AdaMax optimizers, respectively.}
	\label{9}
\end{figure*}

As shown in Figure \ref{9}, the sample distribution representations under different optimizers differ significantly. The Pearson correlation coefficients of SGD - AdaGrad, SGD - AdaMax, and AdaGrad - AdaMax are 0.8914, 0.6877, 0.6962 for the training samples and 0.8877, 0.6628, 0.7062 for the testing samples, respectively. It thus confirms that different optimizers have a great impact on our proposed bi-dimensional representation. Essentially, the optimization algorithm determines the direction of optimization at each step of the training, which leads to changes in the acquisition parameters, resulting in differences in the learning process and thus affecting the representations of the sample distribution.

\section{Application}
\subsection{Training Acceleration}
\begin{figure}
	\centering
	\subfloat[Effect of density calculation parameters $r$ on training performance]{\includegraphics[width=0.5\linewidth]{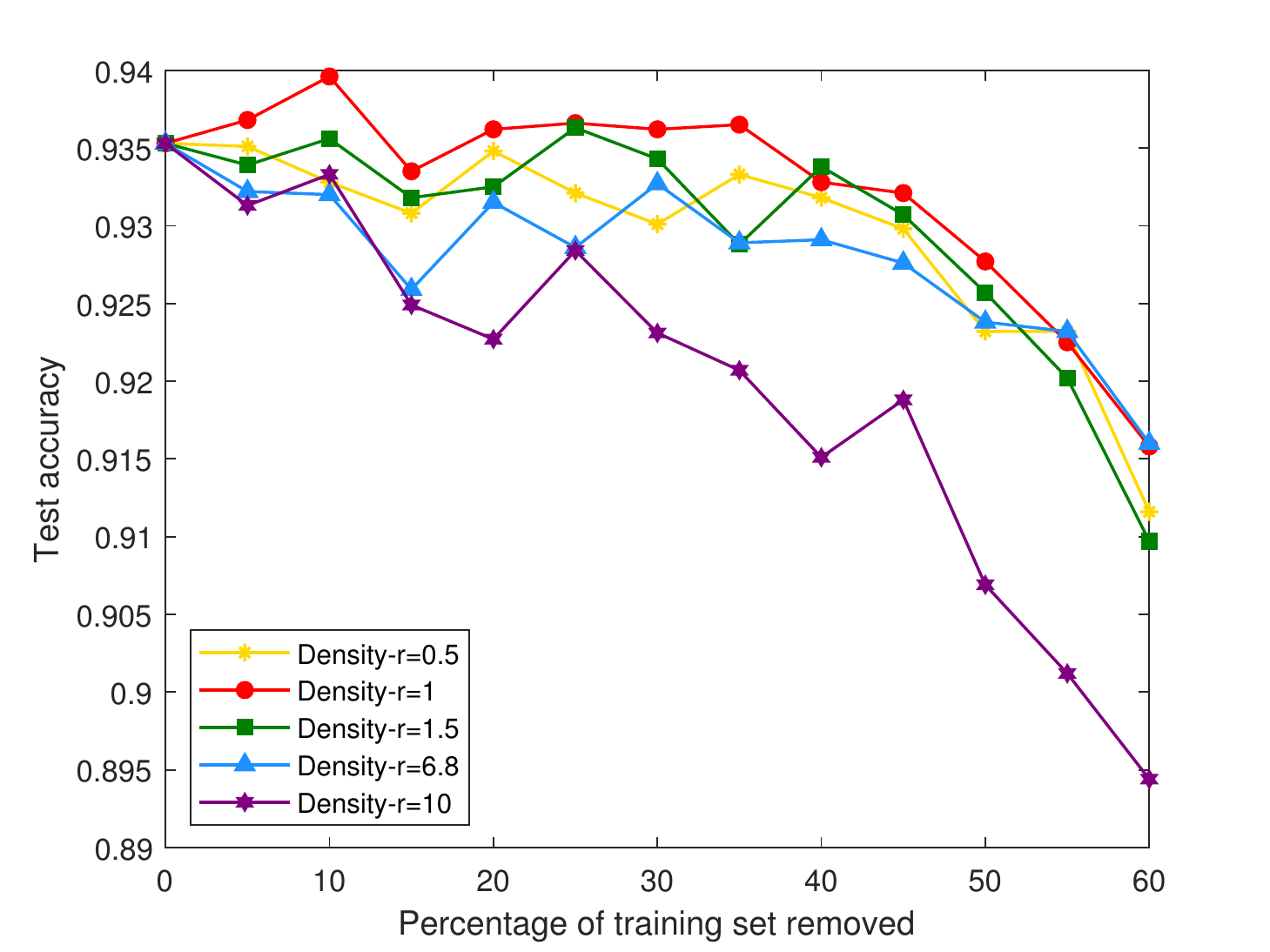}%
		\label{10a}}
	\hfil
	\subfloat[Comparison of different strategies]{\includegraphics[width=0.5\linewidth]{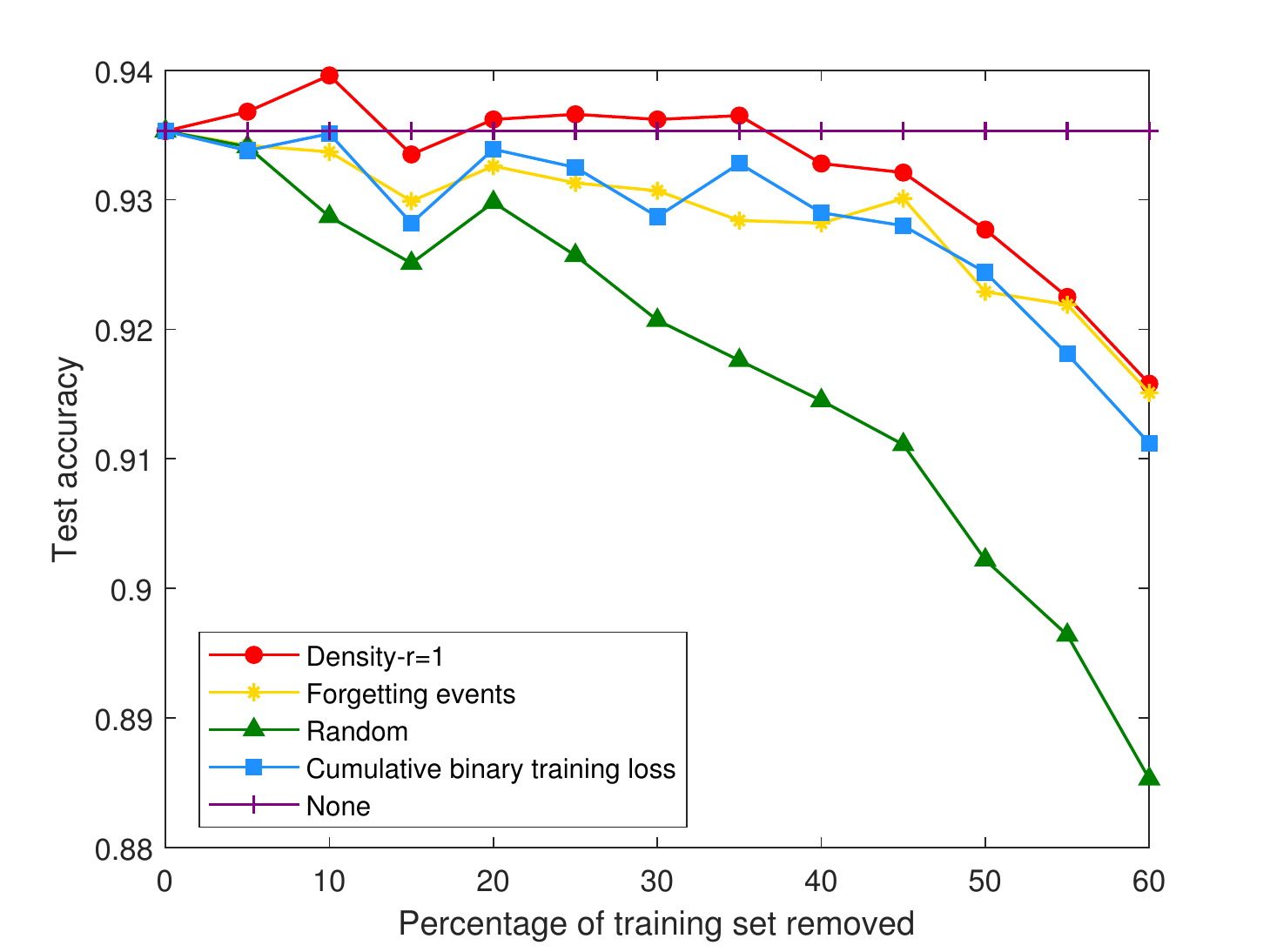}%
		\label{10b}}
	\hfil	
	\caption{Generalization performance on Cifar-10 of ResNet110, in which progressively increasing the proportion of the removed training set.}
	\label{10}
\end{figure}

The Cifar-10 training set is extremely unevenly distributed in our bi-dimensional representation space. The samples in the lower right corner are relatively simple and have little impact on the performance of the final model, but are densely distributed and dominate in the training set. As a result, there are a large number of redundant pattern approximation samples in the training set. We attempt to accomplish an efficient training process with small sample sets by eliminating the high-density redundant samples. We compute the density values of each sample in our bi-dimensional representation space according to the caption of Figure \ref{2} and remove the high-density training sample by sorting them in descending order of density values to investigate the effect of the proportion of removed samples on the generalization performance. Since the computed density value is related to the radius $r$ described in the caption of Figure \ref{2}, we firstly explore the effect of $r$ on training performance as shown in Figure \ref{10} (a). From Figure \ref{10} (a), we can see that $r$ has a large effect on the training performance and the overall performance is relatively best for $r=1$. Therefore, we take the calculated value of density with $r = 1$ for the subsequent experiments. Then, we follow the strategy of removing samples according to the metric of CBTL proposed in Jiang's work\citeup{cscore} and forgetting events proposed in Toneva's work\citeup{forgettingevents} as a comparison with our method, as shown in Figure \ref{10} (b).
It can be observed in Figure \ref{10} (b) that the performance fast drops when the samples are removed randomly. However, when samples are removed according to some strategies, there is no significant effect on the generalization performance, and the test accuracy remains above 0.91 even after removing $60\%$ of the training samples. In addition, the strategy proposed in this paper slightly outperforms the strategies of forgetting events and CBTL. This validates to some extent the insight of dataset compressibility and contributes to accelerating training with small sample set and low time and computational cost.

\subsection{Testing Acceleration}
Unlike the training, which is a pattern extraction process that can reduce sample complexity by removing samples with close pattern representations, the goal of the testing process is often to maximize discrimination of testees' generalization level using minimal sample complexity. This requires reducing the number of samples involved in evaluation while ensuring adequate discrimination for the generalization performance of the testees. It means that more attention needs to be paid to difficult samples in practical experience, which is often achieved by means of difficulty-based sampling. Our proposed bi-dimensional sample representation can be used as a measure of difficulty, obtained by dividing the defined bi-dimensional space based on the sample distribution in a uniform angle. 

As shown by the red dashed line in the Figure \ref{2} (b), we perform uniform difficulty division of the samples by rays emanating from a uniform angle at the center, which is the midpoint of the range of 10,000 testing sample representation's horizontal coordinates values. The smaller the angle is, the finer the difficulty division is. (Note that the figure is only for illustration, please temporarily ignore the problem of inconsistent horizontal and vertical coordinates ratio of the figure.)

We first use the original 10,000 testing samples to evaluate 17 algorithms including deep learning methods and traditional machine learning methods. Then, based on the balanced difficulty sampling method described above, the representation was divided into 4 bins and 10 bins at 45 degrees and 18 degrees, respectively. We randomly sampled the samples in each bin to obtain a smaller dataset containing 400 testing samples, again perform evaluation for the 17 algorithms, and the results are presented in table \ref{table2}.
\begin{table}
	\caption{Test results for (sampled) testing samples}
	\label{table2}
	\centering
	\begin{tabular}{llll}
		\hline\noalign{\smallskip}
		\multirow{2}*{Algorithm} & Original  & 400 samples  & 400 samples \\
		~&10000 samples&(10 bins)&(4 bins)\\
		\noalign{\smallskip}\hline\noalign{\smallskip}
		LeNet    & 0.7505&0.3875&0.5325\\
		NetinNet & 0.8859&0.5125&0.6400\\
		VGG-19 & 0.9338&0.6725&0.7475\\
		ResNet-20  &0.9138&0.5850&0.7050\\
		ResNet-32 & 0.9219&0.5925&0.6850\\
		ResNet-110 &0.9335&0.6175&0.7275\\
		WideRes16$\times8$ &0.9504&0.7450&0.7775\\
		WideRes28$\times10$ &0.9542&0.8150&0.8425\\
		EfficientNet(B8) &0.7353&0.4025&0.4925\\
		DenseNet100$\times12$  &0.9465&0.7825&0.8225\\
		DenseNet160$\times24$ &0.9528&0.7800&0.7925\\
		SVM   &0.5441&0.3075&0.4225\\
		Decision Tree &0.2691&0.1575&0.2475\\
		Logistic Regression &0.3983&0.2800&0.3625\\
		KNN   &0.3398&0.2800&0.3075\\
		MLP  &0.4440&0.3050&0.3925\\
		Random Forest &0.4749&0.3400&0.3875\\
		\noalign{\smallskip}\hline						
	\end{tabular}
\end{table}

It is clear that the algorithm performance drop significantly when tested with samples from balanced difficulty sampling. Especially, it applies to traditional machine learning methods as well. The long-tailed nature of the distribution of the original dataset results in a very low proportion of difficult samples, which are overwhelmed by a large number of simple samples. We increased the proportion of difficult samples by balanced difficulty sampling. The algorithm has worse performance when tested on the samples selected based on 10 bins than 4 bins. This is because the uneven distribution of the original dataset causes that more fine-grained sampling can increase the proportion of difficult samples. We use the test results with original 10000 testing samples as the benchmark. To measure the discriminative ability of our sampled small dataset for the algorithm performance, we calculated the Spearman correlation coefficients of the evaluation results with sampled small datasets based on 10 bins and 4 bins with the benchmark. They are 0.9871 and 0.9877, respectively, both of which have extremely strong correlations and sufficient discriminative ability for the algorithm.

Further, we explore how many samples in each bin sampled on the basis of sampling from 10 bins can most efficiently maintain the discriminative ability of the algorithm performance. We consider the samples on the horizontal axis, i.e., the samples whose forgetting events is 0.

There are only three samples located on the left half-axis, and we sample all of them, followed by $(0^{\circ},18^{\circ})$ for the second bin, $[18^{\circ}, 36^{\circ})$ for the third bin, $\cdots$, $[162^{\circ}, 180^{\circ})$ for the 11th bin, and the $180^{\circ}$ for the last bin. Except for the first bin having only three samples which are all sampled, in each of the remaining bins, we select $n$ samples. 

\begin{figure}
	\centering
	\includegraphics[width=0.5\textwidth]{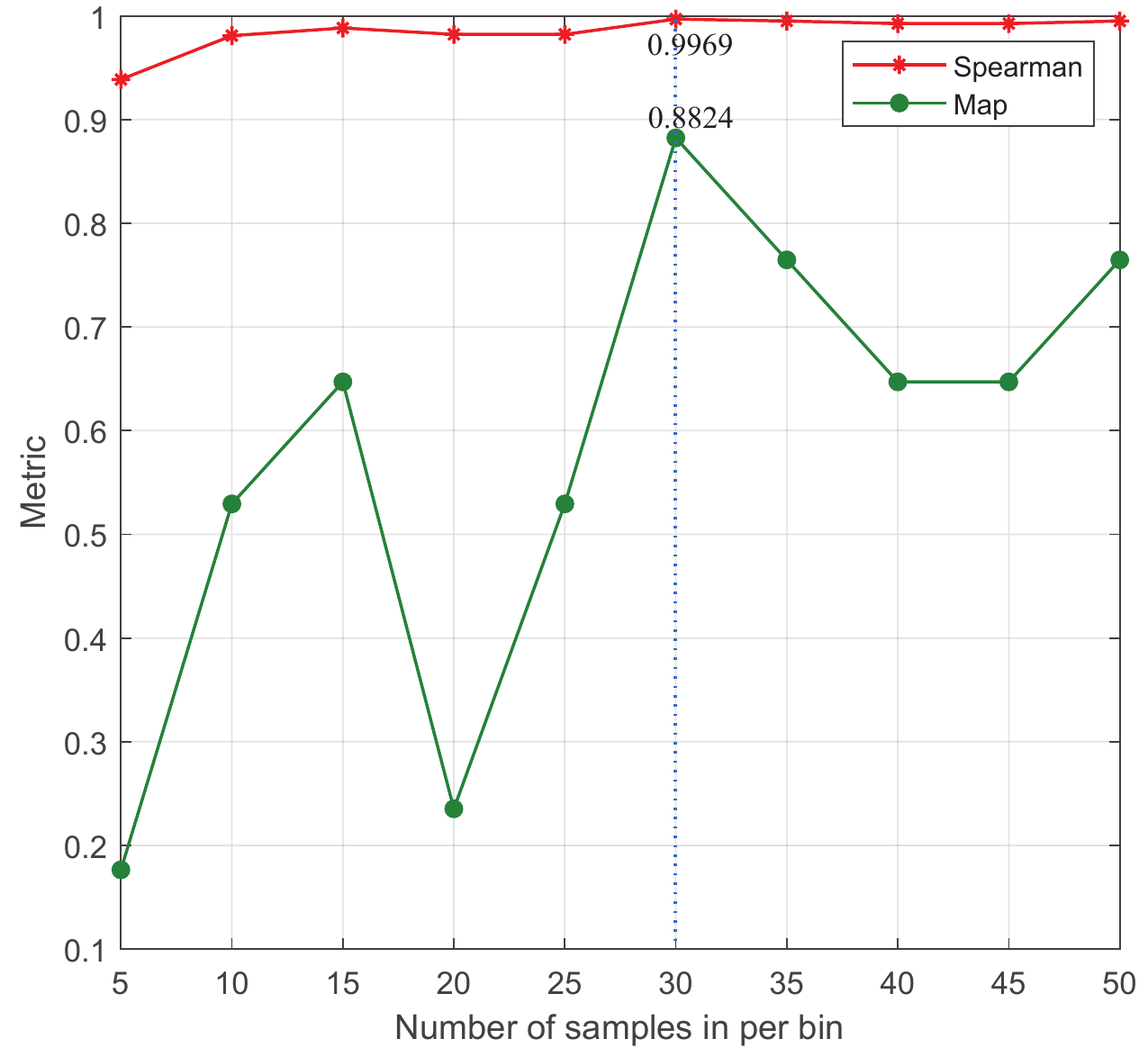}
	\caption{The effect of the number of testing samples sampled in per bin on the algorithm performance evaluation.}
	\label{11}
\end{figure}

Since the least number of samples in the latter 11 bins is 49, we explore the effect of $n$, where $n\leq49$, on the test results. As we are concerned about the discriminative ability of the testing set on the algorithm performance, i.e., more concerned about the ranking stability of the algorithm performance evaluation, we calculate the Spearman correlation coefficient of testing results with $11n+3$ sampled samples compared to that with original 10000 testing samples and the mean average precision of the algorithm ranking. The results are plotted as a line graph as shown in Figure \ref{11}. We can see that sampling 30 samples in each bin, i.e. $11\times30+3=333$ samples, is the best, thus we construct small testing set, Cifar-10-333, to perform fast testing of algorithm performance.

Experiments show that uniform regularity sampling for the testing set can significantly reduce the sample complexity while maintaining discriminative ability on the generalization performance of the tested algorithms.

\section{Conclusion}
In this paper, we propose to measure the sample-wise regularity that certain sample shows when being learned or generalized. We show that for \textit{i.i.d.} training and testing sets, sample distributions are alike under statistics of similar definition, i.e., CBTL/CBGL and forgetting/mal-generalizing events. Inspired by this result, we propose a pair of bi-dimensional representation for measuring sample regularity in network learning and generalization process, respectively. In the property investigation, we find that the proposed measures seem to be fairly stable with respect to the different characteristics of training and testing stage. Further applications in training and testing acceleration show that samples with higher regularity seem to contribute little in both process, which in return validated the effectiveness of the proposed measures.
	
	\section*{Declarations}
	\subsection*{Funding}
	This work was supported by the National Natural
	Science Foundation of China under Grant 61973245.
	\subsection*{Conflicts of interest/Competing interests}
	The authors declare that they have no conflict of interest/competing interests.
	\subsection*{Availability of data and material}
	Not available.
	\subsection*{Author's contributions}
	Chi Zhang conceived the conception and the method of the study; Chi Zhang and Yu Liu jointly designed the experiments and analyzed experimental results (each contributing 42.5\% of the whole research); Yu Liu performed the experiments; Le Wang, Jingxue Hu and Yuehu Liu helped perform the analysis with constructive discussions (each contributing 5\% of the whole research); Chi Zhang and Yu Liu prepared the manuscript, and all authors provided feedback during the manuscript revisions.
	
	\subsection*{Ethics approval}
	Not applicable.
	\subsection*{Consent to participate}
	Not applicable.
	\subsection*{Consent to publication}
	Not applicable.


%
%

\bibliographystyle{spbasic_unsort}      
\bibliography{template}   

%
%

\end{document}